\documentclass[twoside,11pt]{article}
\RequirePackage{fix-cm}
\usepackage{jmlr2e}
\usepackage{units}

\let\jmlrproof\proof
\let\jmlrendproof\endproof
\let\proof\relax
\let\endproof\relax
\usepackage{amsmath,amsthm}
\let\proof\jmlrproof
\let\endproof\jmlrendproof

\usepackage{enumitem}
\usepackage{array}
\usepackage{booktabs}
\usepackage[usenames,dvipsnames]{xcolor}
\usepackage{verbatim}
\usepackage{subfig}
\usepackage[unicode=true,bookmarks=true,bookmarksnumbered=false,bookmarksopen=false, breaklinks=true,backref=false,citecolor=blue,colorlinks=true]{hyperref}
\usepackage[ruled,titlenotnumbered]{algorithm2e}

\providecommand{\tabularnewline}{\\}

\theoremstyle{plain}
\newtheorem{thm}{Theorem}
\newtheorem{lem}{Lemma}
\newtheorem{pro}{Proposition}
\newtheorem{cor}{Corollary}

\theoremstyle{definition}
\newtheorem{dfn}{Definition}
\newtheorem*{asmp*}{Assumption}
\newtheorem{exm}{Example}

\theoremstyle{remark}
\newtheorem{rem}{Remark}

\newcommand{\SCond}{SRH} 
\newcommand{\NSCond}{SRL} 

\newcommand{\Breg}[3]{\ensuremath{\mathrm{B}_{#1}\left(#2\parallel #3\right)}}
\newcommand{\bsym}[1]{\ensuremath{\boldsymbol{#1}}}
\newcommand{\mx}[1]{\ensuremath{\mathbf{#1}}}
\newcommand{\vc}[1]{\ensuremath{\mathbf{#1}}}
\newcommand{\st}[1]{\ensuremath{\mathcal{#1}}}
\newcommand{\norm}[1]{\ensuremath{\left\Vert #1\right\Vert}}
\newcommand{\supp}{\ensuremath{\text{supp}}}

\newcommand{\SBedit}[2][]{{\bf #1} #2} 

\begin{document}

\title{Greedy Sparsity-Constrained Optimization}

\author{           
       \name Sohail Bahmani \email sbahmani@cmu.edu \\
		\addr Department of Electrical and Computer Engineering \\
		Carnegie Mellon University \\
		Pittsburgh, PA 15213, USA
		\AND
		\name Bhiksha Raj \email bhiksha@cs.cmu.edu \\
		\addr Language Technologies Institute \\
		Carnegie Mellon University \\
		Pittsburgh, PA 15213, USA
		\AND
		\name Petros Boufounos \email petrosb@merl.com \\
		\addr Mitsubishi Electric Research Laboratories\\
		Boston, MA 02139, USA}

\editor{}
		
\maketitle

\begin{abstract}%
Sparsity-constrained optimization has wide applicability in machine
learning, statistics, and signal processing problems such as feature selection and Compressive Sensing. A vast body of work has studied the sparsity-constrained optimization from theoretical, algorithmic, and application aspects in the context of sparse estimation in linear models where the fidelity of the estimate is measured by the squared error. In contrast, relatively less effort has been made in the study of sparsity-constrained optimization in cases where nonlinear models are involved or the cost function is not quadratic. In this paper we propose a greedy algorithm, Gradient Support Pursuit (GraSP), to approximate sparse minima of cost functions of arbitrary form. Should a cost function have a Stable Restricted Hessian (SRH) or a Stable Restricted Linearization (SRL), both of which are introduced in this paper, our algorithm is guaranteed to produce a sparse vector within a bounded distance from the true sparse optimum. Our approach generalizes known results for quadratic cost functions that arise in sparse linear regression and Compressive Sensing. We also evaluate the performance of GraSP through numerical simulations on synthetic data, where the algorithm is employed for sparse logistic regression with and without $\ell_2$-regularization.
\end{abstract}
\section{Introduction}
The demand for high-dimensional data analysis has grown significantly over the past decade by the emergence of applications such as social networking, bioinformatics, and mathematical finance. In these applications data samples often have thousands of features using which an underlying parameter must be inferred or predicted. In many circumstances the number of collected samples is significantly smaller than the dimensionality of the data, rendering any inference from the data ill-posed. However, it is widely acknowledged that the data sets that need to be processed usually exhibit significant structure, which sparsity models are often able to capture. This structure can be exploited for robust regression and hypothesis testing, model reduction and variable selection, and more efficient signal acquisition in \emph{underdetermined} regimes. Estimation of parameters with sparse structure is usually cast as an optimization problem, formulated according to specific application requirements. Developing techniques that are robust and computationally tractable to solve these optimization problems, even only  approximately, is therefore critical.

In particular, theoretical and application aspects of sparse estimation in linear models have been studied extensively in areas such as signal processing, machine learning, and statistics. However, sparse estimation in problems where nonlinear models are involved have received comparatively little attention. Most of the work in this area extend the use of the $\ell_1$-norm as a regularizer, effective to induce sparse solutions in linear regression, to problems with nonlinear models \citep[see e.g.,][]{Bunea08,vanDeGeer08,kakade10learning,Negahban-M-estimators}. As a special case, logistic regression with $\ell_1$ and elastic net regularization are studied by \citet{Bunea08}. Furthermore, \citet{kakade10learning} have studied the accuracy of sparse estimation through $\ell_1$-regularization for the exponential family distributions. A more general frame of study is proposed and analyzed by \citet{Negahban-M-estimators} where regularization with ``decomposable'' norms is considered in M-estimation problems. To provide the accuracy guarantees,  these works generalize the Restricted Eigenvalue condition \citep{BickelRitovTsybakov09} to ensure that the loss function is strongly convex over a restriction of its domain. We would like to emphasize that these sufficient conditions generally hold with proper constants and with high probability only if one assumes that the true parameter is bounded. This fact is more apparent in some of the mentioned work \citep[e.g.,][]{Bunea08,kakade10learning}, while in some others \citep[e.g.,][]{Negahban-M-estimators} the assumption is not explicitly stated. We will elaborate on this matter in \S\ref{sec:bg}. \cite{tewari_greedy_2011} also proposed a coordinate-descent type algorithm for minimization of a convex and smooth objective over the convex signal/parameter models introduced in \citep{chandrasekaran_convex_2010}. This formulation includes the $\ell_1$-constrained minimization as a special case, and the algorithm is shown to converge to the minimum in objective value similar to the standard results in convex optimization. 

Furthermore, \cite{Trad_TZ10} proposed a number of greedy that
sparsify a given estimate at the cost of relatively small increase of
the objective function. However, their algorithms are not
stand-alone. A generalization of Compressed Sensing is also proposed
in \citep{NLCS_Blumensath}, where the linear measurement operator is
replaced by a nonlinear operator that applies to the sparse
signal. Considering the norm of the residual error as the objective,
\cite{NLCS_Blumensath} shows that if the objective satisfies certain
sufficient conditions, the sparse signal can be accurately estimated
by a generalization of the Iterative Hard Thresholding algorithm
\citep{blumensath_iterative_2009}. The formulation of
\citep{NLCS_Blumensath}, however, has a limited scope because the
metric of error is defined using a norm. For instance, the formulation
does not apply to objectives such as the logistic loss. More recently, \cite{jalali_learning_2011}
  studied a forward-backward algorithm using a variant of the
  sufficient conditions introduced in \citep{Negahban-M-estimators}.
  Similar to our work, the main result in \citep{jalali_learning_2011}
  imposes conditions on the function as restricted to sparse inputs
  whose non-zeros are fewer than a multiple of the target sparsity
  level. The multiplier used in their results has an \emph{objective-dependent} value and is never less than 10. Furthermore, the multiplier is important in their analysis not only for determining the stopping condition of the algorithm, but also in the lower bound assumed for the minimal magnitude of the non-zero entries. In contrast, the multiplier in our results is fixed at 4, independent of the objective function itself, and we make no assumptions
about the magnitudes of the non-zero entries.

This paper presents an extended version with improved
  guarantees of our prior work in \citep{GraSP-BBR11}, where we
  proposed a greedy algorithm, the Gradient Support Pursuit (GraSP),
  for sparse estimation problems that arise in applications with
  general nonlinear models. We prove the accuracy of GraSP for a
class of cost functions that have a \emph{Stable Restricted Hessian}
(\SCond{}). The \SCond{}, introduced in \citep{GraSP-BBR11},
characterizes the functions whose restriction to sparse canonical
subspaces have well-conditioned Hessian matrices. Similarly, we
analyze the GraSP algorithm for non-smooth functions that have a
\emph{Stable Restricted Linearization} (\NSCond{}), a property
introduced in this paper, analogous to \SCond{}. The analysis and the
guarantees for smooth and non-smooth cost functions are similar,
except for less stringent conditions derived for smooth cost functions
due to properties of symmetric Hessian matrices. We also
  prove that the \SCond{} holds for the case of the $\ell_2$-penalized
  logistic loss function.

\subsection*{Notation} In the remainder of this paper we use the notation listed in Table \ref{tab:Notation}.

\begin{table}
\caption{Notation used in this paper\label{tab:Notation}}
\centering%
\begin{tabular*}{1.0\columnwidth}{@{\extracolsep{\fill}}>{\raggedright}m{0.1\columnwidth}>{\raggedright}m{0.8\columnwidth}}
\toprule 
\textbf{\large Symbol} & \textbf{\large Description}\tabularnewline
\midrule 
$\left[n\right]$ & the set $\left\{ 1,2,\ldots,n\right\} $ for any $n\in \mathbb{N}$\tabularnewline
\midrule 
$\st I$ & calligraphic letters denote sets unless stated otherwise (e.g., $\st{N}\left(\mu,\sigma^2\right)$ denotes a normal distribution)\tabularnewline
\midrule 
$\st{I}^c$ & complement of set $\st{I}$ \tabularnewline 
\midrule 
$\vc v$ & bold face small letters denote column vectors in $\mathbb{R}^{b}$
for some $b\in\mathbb{N}$\tabularnewline
\midrule 
$\norm{\vc v}_{q}$ & the $\ell_{q}$-norm of vector $\vc v$, that is $\left(\sum_{i=1}^{b}\left|v_{i}\right|^{q}\right)^{1/q}$,
for a real number $q\geq1$\tabularnewline
\midrule 
$\norm{\vc v}_{0}$ & the ``$\ell_{0}$-norm'' of vector $\vc v$ that merely counts its
nonzero entries\tabularnewline
\midrule 
$\vc v|_{\st I}$ & depending on the context
\parbox{1\columnwidth}{%
 \begin{enumerate}[topsep = 0pt, parsep = 0pt, itemsep = 0pt]
	\item restriction of vector $\vc v$ to the rows indicated by indices in $\st I$, or
	\item a vector that equals $\vc{v}$ except for coordinates in $\st{I}^c$ where it is zero
\end{enumerate}%
}\tabularnewline\midrule 
$\vc v_{r}$ & the best $r$-term approximation of vector $\vc v$\tabularnewline
\midrule 
$\supp\left(\vc v\right)$ & the support set (i.e., indices of the non-zero entries) of $\vc v$\tabularnewline
\midrule 
$\mx M$ & bold face capital letters denote matrices in $\mathbb{R}^{a\times b}$
for some $a,b\in\mathbb{N}$\tabularnewline
\midrule 
$\mx M^{\mathrm{T}}$ & transpose of matrix $\mx M$\tabularnewline
\midrule 
$\mx M^{\dagger}$ & pseudo-inverse of matrix $\mx M$\tabularnewline
\midrule 
$\mx M_{\st I}$ & restriction of matrix $\mx M$ to the columns enumerated by $\st I$\tabularnewline
\midrule 
$\norm{\mx{M}}$ & the operator norm of matrix $\mx{M}$ which is equal to $\sqrt{\lambda_{\max}\left(\mx{M}^\mathrm{T}\mx{M}\right)}$\tabularnewline
\midrule
$\mx{I}$ & the identity matrix\tabularnewline
\midrule
$\mx{P}_\st{I}$ & restriction of the identity matrix to the columns indicated by $\st{I}$\tabularnewline
\midrule
$\vc{1}$ & column vector of all ones\tabularnewline
\midrule
$\mathbb{E} \left[\cdot\right]$ & expectation\tabularnewline
\midrule
$\mx{H}_f\left(\cdot\right)$ & Hessian of the function $f$\tabularnewline
\bottomrule
\end{tabular*}
\end{table}

\subsection*{Paper Outline} In \S\ref{sec:bg} we provide a background on sparse parameter estimation which serves as an overview of prior work. In \S\ref{sec:algorithm}
we state the general formulation of the problem and present our algorithm. Conditions that characterize the cost functions and the main accuracy guarantees of our algorithm are provided in \S\ref{sec:algorithm} as well. The guarantees of the algorithm are proved in Appendices \ref{app:Smooth} and \ref{app:Nonsmooth}. As an example where our algorithm can be applied, $\ell_2$-regularized logistic regression is studied in \S\ref{sec:l2-log}. Some experimental results for logistic regression with sparsity constraints are presented in \S\ref{sec:experiments}. Finally, \S\ref{sec:conclusion} discusses the results and concludes.

\section{Background}\label{sec:bg}
We first briefly review sparse estimation problems studied in the literature.
\subsection{Sparse Linear Regression and Compressed Sensing}\label{ssec:cs}
The special case of sparse estimation in linear models has gained significant attention under the title of Compressed Sensing (CS)~\citep{donoho_compressed_2006}. In standard CS problems the aim is to estimate a sparse vector $\vc{x}^\star$ from noisy linear measurements $\vc{y}=\mx{A}\vc{x}^\star+\vc{e}$, where $\mx{A}$ is a known $n\times p$ measurement matrix with $n\ll p$ and $\vc{e}$ is the additive measurement noise. To find the sparsest estimate in this \emph{underdetermined} problem that is consistent with the measurements $\vc{y}$ one needs to solve the optimization problem
\begin{align} 
\widehat{\vc{x}}=&\arg\min_{\vc{x}} \norm{\vc{x}}_{0}\quad\text{s.t. }
\norm{\vc{y}-\mx{A}\vc{x}}_{2}\leq\varepsilon,\label{eq:E1}
\end{align} 
 where $\varepsilon$ is a given upper bound for $\norm{\vc{e}}_2$~\citep{CandesRombergTao06}. In the absence of noise (i.e., when $\varepsilon =0$), if $\vc{x}^\star$ is $s$-sparse (i.e., it has at most $s$ nonzero entries) one merely needs every $2s$ columns of $\mx{A}$ to be linearly independent to guarantee exact recovery~\citep{DonohoElad2003optimally}. Unfortunately, the ideal solver (\ref{eq:E1}) is computationally NP-hard in general~\citep{Natarajan1995sparse} and one must seek approximate solvers instead.

It is shown in~\citep{CandesRombergTao06} that under certain conditions, minimizing the $\ell_{1}$-norm as a convex proxy for the $\ell_{0}$-norm yields accurate estimates of $\vc{x}^\star$. The resulting approximate solver basically returns the solution to the convex optimization problem 
\begin{align} 
\widehat{\vc{x}}=&\arg\min_{\vc{x}} \norm{\vc{x}}_{1}\quad\text{s.t. }
\norm{\vc{y}-\mx{A}\vc{x}}_{2}\leq\varepsilon,\label{eq:E2}
\end{align}
The required conditions for approximate equivalence of (\ref{eq:E1}) and (\ref{eq:E2}), however, generally hold only if measurements are collected at a higher rate. Ideally, one merely needs $n=O\left(s\right)$ measurements to estimate $\vc{x}^\star$, but $n = O(s\log \nicefrac{p}{s})$ measurements are necessary for the accuracy of (\ref{eq:E2}) to be guaranteed.

The convex program (\ref{eq:E2}) can be solved in polynomial time using interior point methods. However, these methods do not scale well as the size of the problem grows. Therefore, several first-order convex optimization methods are developed and analyzed as more efficient alternatives (see e.g., \cite{beck2009FISTA} and \cite{AgarwalPGD}). Another category of low-complexity algorithms in CS are the non-convex \emph{greedy pursuits} including Orthogonal Matching Pursuit (OMP)
\citep{PatiRezaiifarKrishnaprasad93,tropp_signal_2007}, Compressive Sampling Matching
Pursuit (CoSaMP) \citep{NeedellTropp09}, Iterative Hard Thresholding
(IHT) \citep{blumensath_iterative_2009}, and Subspace Pursuit
\citep{wei_dai_subspace_2009} to name a few. These greedy algorithms implicitly approximate the solution to the $\ell_0$-constrained least squares problem
\begin{align}
\widehat{\vc{x}}=&\arg\min_{\vc{x}} \frac{1}{2}\norm{\vc{y}-\mx{A}\vc{x}}_{2}^{2}\quad\text{s.t. }
\norm{\vc{x}}_{0}\leq s.\label{eq:E3}
\end{align}
The main theme of these iterative algorithms is to use the residual error from the previous iteration to successively approximate the position of non-zero entries and estimate their values. These algorithms have shown to exhibit accuracy guarantees similar to those of convex optimization methods, though with more stringent requirements.

As mentioned above, to guarantee accuracy of the CS algorithms the measurement matrix should meet certain conditions such as \emph{incoherence} \citep{DonohoHuo01}, Restricted Isometry Property (RIP)~\citep{CandesRombergTao06}, Nullspace Property~\citep{Cohen2009NSP}, etc. Among these conditions RIP is the most commonly used and the best understood condition. 

Matrix $\mx{A}$ is said to satisfy the RIP of order $k$---in its symmetric form---with constant $\delta_{k}$, if $\delta_{k} < 1$ is the smallest number that
\begin{align*}
\left(1-\delta_k\right)\norm{\vc{x}}_2^2\leq\norm{\mx{A}\vc{x}}_2^2\leq\left(1+\delta_k\right)\norm{\vc{x}}_2^2
\end{align*}
 holds for all $k$-sparse vectors $\vc{x}$. Several CS algorithms are shown to produce accurate solutions provided that the measurement matrix has a sufficiently small RIP constant of order $ck$ with $c$ being a small integer. For example, solving \eqref{eq:E2} is guaranteed to yield an accurate estimate of $s$-sparse $\vc{x}^\star$ if $\delta_{2s}<\sqrt{2}-1$~\citep{Candes2008589}. Interested readers can find the best known RIP-based accuracy guarantees for some of the CS algorithms in~\citep{RIP-Foucart2010}.

\subsection{Beyond Linear Models}
The CS reconstruction algorithms attempt to provide a sparse vector that incurs only a small squared error which measures consistency of the solution versus the acquired data. While
this measure of discrepancy is often desirable for signal processing applications, it is not
the appropriate choice for a variety of other applications. For
example, in statistics and machine learning the logistic loss function
is also commonly used in regression and classification problems
\citep[see][and references therein]{LASPLORE}. Thus, it is desirable to
develop theory and algorithms that apply to a broader class of optimization problems with sparsity constraints.

 The existing studies on this subject are mostly in the context of statistical estimation. The majority of these studies consider the cost function to be convex everywhere and rely on the $\ell_{1}$-regularization as the means to induce sparsity in the solution. For example, \citet{kakade10learning} have shown that for the exponential family of distributions maximum likelihood estimation with $\ell_{1}$-regularization yields accurate estimates of the underlying sparse parameter. Furthermore, Negahban et al. have developed a unifying framework for analyzing statistical accuracy of \emph{$M$-estimators} regularized by ``decomposable'' norms in~\citep{Negahban-M-estimators}. In particular,  in their work $\ell_{1}$-regularization is applied to Generalized Linear Models (GLM)~\citep{dobson_introduction_2008} and shown to guarantee a bounded distance between the estimate and the true statistical parameter. To establish this error bound they introduced the notion of \emph{Restricted Strong Convexity} (RSC), which basically requires a lower bound on the curvature of the cost function around the true parameter in a restricted set of directions. The achieved error bound in this framework is inversely proportional to this curvature bound. Furthermore, \citet{AgarwalPGD} have studied Projected Gradient Descent as a method to solve $\ell_1$-constrained optimization problems and established accuracy guarantees using a slightly different notion of RSC and \emph{Restricted Smoothness} (RSM).
 
 Note that the guarantees provided for majority of the $\ell_1$-regularization algorithms presume that the true parameter is bounded, albeit implicitly. For instance, the error bound for $\ell_1$-regularized logistic regression is recognized by \citet{Bunea08} to be dependent on the true parameter \citep[Assumption A, Theorem 2.4, and the remark that succeeds them]{Bunea08}. Moreover, the result proposed by \citet{kakade10learning} implicitly requires the true parameter to have a sufficiently short length  to allow the choice of the desirable regularization coefficient \citep[Theorems 4.2 and 4.5]{kakade10learning}. \citet{Negahban-M-estimators} also assume that the true parameter is inside the unit ball to establish the required condition for their analysis of $\ell_1$-regularized GLM, although this restriction is not explicitly stated \citep[see the longer version of][p. 37]{Negahban-M-estimators}. We can better understand why restricting the length of the true parameter may generally be inevitable by viewing these estimation problems from the perspective of empirical processes and their convergence. The empirical processes, including those considered in the studies mentioned above, are generally good approximations of their corresponding expected process (see \citep[Chapter 5]{Vapnik98} and \citep{VandeGeer00}). Therefore, if the expected process is not strongly convex over an unbounded, but perhaps otherwise restricted, set the corresponding empirical process cannot be strongly convex over the same set. This reasoning applies in many cases including the studies mentioned above, where it would be impossible to achieve the desired restricted strong convexity properties---with high probability---if the true parameter is allowed to be unbounded. 

Furthermore, the methods that rely on the $\ell_{1}$-norm are known to result in sparse solutions, but, as mentioned in~\citep{kakade10learning}, the sparsity of these solutions is not known to be optimal in general. One can intuit this fact from definitions of RSC and RSM. These two properties bound the curvature of the function from below and above in a restricted set of directions around the true optimum. For quadratic cost functions, such as squared error, these curvature bounds are absolute constants. As stated before, for more general cost functions such as the loss functions in GLMs, however, these constants will depend on the location of the true optimum. Consequently, depending on the location of the true optimum these error bounds could be extremely large, albeit finite. When error bounds are significantly large, the sparsity of the solution obtained by $\ell_1$-regularization may not be satisfactory. This motivates investigation of algorithms that do not rely on $\ell_{1}$-norm as a proxy.

\section{Problem Formulation and the GraSP Algorithm}\label{sec:algorithm}
As seen in \S\ref{ssec:cs}, in standard CS the squared error
$f(\vc{x})=\frac{1}{2}\norm{\vc{y}-\mx{A}\vc{x}}_2^2$ is used to measure data fidelity. While
this is appropriate for a large number of signal acquisition
applications, it is not the right cost in other fields. Thus, the significant advances in CS cannot readily be applied in
these fields when estimation or prediction of sparse parameters become necessary. In this paper we focus on a generalization of \eqref{eq:E3} where a
generic cost function replaces the squared error. Specifically, for the cost function $f:\mathbb{R}^p\mapsto\mathbb{R}$, it is desirable to approximate
\begin{align} 
  \arg\min_{\vc{x}} f\left(\vc{x}\right)\quad\text{s.t. }
  \norm{\vc{x}}_{0}\leq s\label{eq:gen_prob}.
\end{align}
We propose the Gradient Support Pursuit (GraSP) algorithm, which is inspired by and generalizes the CoSaMP algorithm, to approximate the solution to (\ref{eq:gen_prob}) for a broader class of cost functions.

Of course, even for a simple quadratic objective,~\eqref{eq:gen_prob}
can have combinatorial complexity and become NP-hard. However, similar to the results of CS,  knowing that the cost function obeys certain properties allows us to obtain accurate estimates through tractable algorithms. \SBedit{To guarantee that GraSP yields accurate solutions and is a tractable algorithm, we also require the cost function to have certain properties that will be described in \S\ref{ssec:conditions}.} These properties are analogous to and generalize the RIP in the standard CS framework. For smooth cost functions we introduce the notion of a Stable Restricted Hessian (\SCond{}) and for non-smooth cost functions we introduce the Stable Restricted Linearization (\NSCond{}). Both of these properties basically bound the Bregman divergence of the cost function restricted to sparse canonical subspaces. However, the analysis based on the \SCond{} is facilitated by matrix algebra that results in somewhat less restrictive requirements for the cost function.

\subsection{Algorithm Description} \label{ssec:AlgDescription}
\begin{algorithm}
\DontPrintSemicolon
\SetKwInOut{Input}{input}\SetKwInOut{Output}{output}
\Input{$f\left(\cdot\right)$ and $s$}
\Output{$\hat{\vc{x}}$}
\BlankLine
{\bf initialize: $\widehat{\vc{x}}=0$}\;
\Repeat{halting condition holds}{
{\bf compute local gradient:} $\vc{z}=\nabla f\left(\widehat{\vc{x}}\right)$\;
{\bf identify directions:} $\st{Z}=\text{supp}\left(\vc{z}_{2s}\right)$\;
{\bf merge supports:} $\st{T}=\st{Z}\cup\text{supp}\left(\widehat{\vc{x}}\right)$\;
{\bf minimize over support:} $\vc{b}=\arg\min f\left(\vc{x}\right) \text{ s.t. } \left.\vc{x}\right\vert_{\st{T}^c}=\vc{0}$\;
{\bf prune estimate:} $\widehat{\vc{x}} = \vc{b}_s$\;
}
\caption{The GraSP algorithm}
\label{Algo}
\end{algorithm}

GraSP is an iterative algorithm, summarized in Algorithm~\ref{Algo}, that
maintains and updates an estimate $\widehat{\vc{x}}$ of the sparse
optimum at every iteration.  The first step in each iteration,
$\vc{z}=\nabla f\left(\widehat{\vc{x}}\right)$, evaluates the gradient of
the cost function at the current estimate. For nonsmooth functions, instead of the gradient we use the \emph{restricted subgradient} $\vc{z} = \nabla_{f}\left(\widehat{\vc{x}}\right)$ defined in \S\ref{ssec:conditions}. Then $2s$ coordinates of the vector $\vc{z}$ that have the largest magnitude are chosen as the directions in which
pursuing the minimization will be most effective. Their indices, denoted by $\st{Z}=\supp\left(\vc{z}_{2s}\right)$, are then merged with
the support of the current estimate to obtain
\mbox{$\st{T}=\st{Z}\cup\text{supp}\left(\widehat{\vc{x}}\right)$}. The combined
support is a set of at most $3s$ indices over which the function $f$
is minimized to produce an intermediate estimate $\vc{b}\!=\!\arg\min
f\left(\vc{x}\right) \text{ s.t. } \left.\vc{x}\right\vert_{\st{T}^c}\!=\!0$. The estimate $\widehat{\vc{x}}$ is then updated as the best $s$-term approximation of the intermediate estimate $\vc{b}$. \SBedit{The iterations terminate once certain condition, e.g., on the change of the cost
function or the change of the estimated minimum from the previous
iteration, holds.}

In the special case where the squared error
$f\left(\vc{x}\right)=\frac{1}{2}\norm{\vc{y}-\mx{A}\vc{x}}_2^2$ is the cost function,
GraSP reduces to CoSaMP. Specifically, the gradient step reduces to
the proxy step $\vc{z}=\mx{A}^{\mathrm{T}}\left(\vc{y}-\mx{A}\hat{\vc{x}}\right)$ and minimization over the restricted support reduces to the constrained pseudoinverse step
$\left.\vc{b}\right\vert_{\st{T}}=\mx{A}_{\st{T}}^{\dagger}\vc{y}$, $\left.\vc{b}\right\vert_{\st{T}^{c}}=\vc{0}$ in CoSaMP.
\paragraph*{Variants}
Although in this paper we only analyze the standard form of GraSP outlined in Algorithm \ref{Algo}, other variants of the algorithm can also be studied. \SBedit{Below we list some of these variants.}

\begin{enumerate}
\item \emph{Debiasing}: In this variant, instead of performing a hard thresholding on the vector $\vc{b}$, the objective is minimized restricted to the support set of $\vc{b}_s$ to obtain the new iterate:
\begin{alignat*}{1}
	\widehat{\vc{x}} & = \arg\min_{\vc{x}}\ f\left(\vc{x}\right)\quad\text{s.t. }\text{supp}\left(\vc{x}\right)\subseteq\text{supp}\left(\vc{b}_s\right).
\end{alignat*}
\item \emph{Restricted Newton Step}: To reduce the computations in each iteration, the minimization that yields $\vc{b}$, we can set $	\left.\vc{b}\right\vert_{\st{T}^c} = \vc{0}$ and take a restricted Newton step as
\begin{alignat*}{1}
\left.\vc{b}\right\vert_{\st{T}} &= \left.\widehat{\vc{x}}\right\vert_\st{T}-\kappa\left(\mx{P}^{\mathrm{T}}_\st{T}\mx{H}_f\left(\widehat{\vc{x}}\right)\mx{P}_\st{T}\right)^{-1}\left.\widehat{\vc{x}}\right|_\st{T},
\end{alignat*}
where $\kappa>0$ is a step-size. Of course, here we are assuming that the restricted Hessian, $\mx{P}^{\mathrm{T}}_\st{T}\mx{H}_f\left(\widehat{\vc{x}}\right)\mx{P}_\st{T}$, is invertible.
\item \emph{Restricted Gradient Descent}: The minimization step can be relaxed even further by applying a restricted gradient descent. In this approach, we again set $\left.\vc{b}\right\vert_{\st{T}^c}=\vc{0}$ and 
\begin{alignat*}{1}
	\left.\vc{b}\right\vert_{\st{T}} &= \left.\widehat{\vc{x}}\right\vert_\st{T} - \kappa \left.\nabla f\left(\widehat{\vc{x}}\right)\right\vert_\st{T}.
\end{alignat*}
Since $\st{T}$ contains both the support set of $\widehat{\vc{x}}$ and the $2s$-largest entries of $\nabla f\left(\widehat{\vc{x}}\right)$ , it is easy to show that each iteration of this alternative method is equivalent to a standard gradient descent followed by a hard thresholding. In particular, if the squared error is the cost function as in standard CS, this variant reduces to the IHT algorithm.
\end{enumerate}

\subsection{Sparse Reconstruction Conditions}\label{ssec:conditions}
In what follows we characterize the functions for which accuracy of GraSP can be guaranteed. For twice continuously differentiable functions we rely on Stable Restricted Hessian (\SCond{}), while for non-smooth cost functions we introduce the Stable Restricted Linearization (\NSCond). These properties that are analogous to the RIP in the standard CS framework, basically require that the curvature of the cost function over the sparse subspaces can be bounded locally from above and below such that the corresponding bounds have the same order. Below we provide precise definitions of these two properties.
\begin{dfn}[Stable Restricted Hessian]
\label{def:D1}Suppose that $f$ is a twice continuously differentiable
function whose Hessian is denoted by $\mx{H}_{f}\left(\cdot\right)$. Furthermore,
let \begin{align}
A_{k}\left(\vc{x}\right)=&\sup\left\{ \bsym{\Delta}^{\mathrm{T}}\mx{H}_{f}\left(\vc{x}\right)\bsym{\Delta}\:\Big|\:
\left\vert\mathrm{supp}\left(\vc{x}\right)\cup\mathrm{supp}\left(\bsym{\Delta}\right)\right\vert\leq k, \norm{\bsym{\Delta}}_2 =1\right\}  \label{eq:E00}\\
\intertext{and} B_{k}\left(\vc{x}\right)=&\inf \left\{ \bsym{\Delta}^{\mathrm{T}}\mx{H}_{f}\left(\vc{x}\right)\bsym{\Delta}\:\Big|:
\left\vert\mathrm{supp}\left(\vc{x}\right)\cup\mathrm{supp}\left(\bsym{\Delta}\right)\right\vert\leq k, \norm{\bsym{\Delta}}_2 =1\right\} , \label{eq:E01}\end{align}
 for all $k$-sparse vectors $\vc{x}$. Then $f$ is said to have a Stable Restricted Hessian (\SCond{}) with constant $\mu_{k}$, or in short $\mu_{k}$-\SCond{},
if $1\leq\frac{A_{k}\left(\vc{x}\right)}{B_{k}\left(\vc{x}\right)}\leq\mu_{k}$.\end{dfn}

\begin{rem}
Since the Hessian of $f$ is symmetric, an equivalent for Definition \ref{def:D1}
is that a twice continuously differentiable function $f$ has $\mu_{k}$-\SCond{}
if the condition number of $\mx{P}_{\st{K}}\mx{H}_{f}\left(\vc{x}\right)\mx{P}_{\st{K}}^{\mathrm{T}}$
is not greater than $\mu_{k}$ for all $k$-sparse vectors $\vc{x}$ and sets $\st{K}\subseteq\left[\SBedit{p}\right]$ with $\left\vert\supp\left(\vc{x}\right)\cup\st{K}\right\vert\leq k$.

In the special case when the cost function is the squared error as in
\eqref{eq:E3}, we can write
$\mx{H}_{f}\left(\vc{x}\right)=\mx{A}^\mathrm{T}\mx{A}$ which is constant. The \SCond{}
condition then requires \begin{align*}
  B_k\norm{\bsym{\Delta}}_2^2\leq\norm{\mx{A}\bsym{\Delta}}_2^2\leq A_k\norm{\bsym{\Delta}}_2^2 \end{align*} 
 to hold for all $k$-sparse vectors $\bsym{\Delta}$ with $A_k/B_k\leq\mu_k$. Therefore, in this special case the \SCond{} condition essentially becomes equivalent to the RIP condition.
\end{rem}
\begin{rem}
Note that the functions that satisfy the \SCond{} are convex over canonical sparse subspaces, but they are not necessarily convex everywhere. The following two examples describe some non-convex functions that have \SCond{}.
\begin{exm}
Let $f\left(\vc{x}\right)=\frac{1}{2}\vc{x}^\mathrm{T}\mx{Q}\vc{x}$, where $\mx{Q}=2\times\vc{1}\vc{1}^\mathrm{T}-\mx{I}$. Obviously, we have $\mx{H}_f\left(\vc{x}\right)=\mx{Q}$. Therefore, \eqref{eq:E00} and \eqref{eq:E01} determine the extreme eigenvalues across all of the $k \times k$ symmetric submatrices of $\mx{Q}$. Note that the diagonal entries of $\mx{Q}$ are all equal to one, while its off-diagonal entries are all equal to two. Therefore, for any $1$-sparse signal $\vc{u}$ we have $\vc{u}^\mathrm{T}\mx{Q}\vc{u} = \norm{\vc{u}}_2^2$, meaning that $f$ has $\mu_1$-\SCond{} with $\mu_1=1$. However, for $\vc{u} = \left[1, -1, 0, \ldots, 0\right]^\mathrm{T}$ we have $\vc{u}^\mathrm{T}\mx{Q}\mathrm{u}<0$, which means that the Hessian of $f$ is not positive semi-definite (i.e., $f$ is not convex). 
\end{exm}
\begin{exm}
Let $f\left(\vc{x}\right)=\frac{1}{2}\norm{\vc{x}}_2^2+Cx_1x_2\cdots x_{k+1}$ where the dimensionality of $\vc{x}$ is greater than $k$. It is obvious that this function is convex for $k$-sparse vectors as  $x_1x_2\cdots x_{k+1}=0$ for any $k$-sparse vector. So we can easily verify that $f$ satisfies \SCond{} of order $k$. However, for $x_1=x_2=\cdots=x_{k+1}=t$ and $x_i=0$ for $i>k+1$ the restriction of the Hessian of $f$ to indices in $\left[k+1\right]$ (i.e., $\mx{P}^\mathrm{T}_{\left[k+1\right]}\mx{H}_f\left(\vc{x}\right)\mx{P}_{\left[k+1\right]}$) is a matrix with diagonal entries all equal to one and off-diagonal entries all equal to $Ct^{k-1}$. Let $\mx{Q}$ denote this matrix and $\vc{u}$ be a unit-norm vector such that $\left\langle \vc{u},\vc{1}\right\rangle=0$. Then it is straightforward to verify that $\vc{u}^\mathrm{T}\mx{Q}\vc{u} = 1-Ct^{k-1}$, which can be negative for sufficiently large values of $C$ and $t$. Therefore, the Hessian of $f$ is not positive semi-definite everywhere, meaning that $f$ is not convex.
\end{exm}
\end{rem}

To generalize the notion of \SCond{} to the case of nonsmooth functions, first we define the \emph{restricted subgradient} of a function.
\begin{dfn}[Restricted Subgradient]
 We say vector $\nabla_{f}\left(\vc{x}\right)$ is a restricted subgradient of $f:\mathbb{R}^{p}\mapsto\mathbb{R}$
at point $\vc{x}$ if 
\begin{align*}
f\left(\vc{x}+\bsym{\Delta}\right)-f\left(\vc{x}\right) & \geq\left\langle \nabla_{f}\left(\vc{x}\right),\bsym{\Delta}\right\rangle 
\end{align*}
 holds for all $k$-sparse vectors $\bsym{\Delta}$.\end{dfn}
  
\begin{rem}
We introduced the notion of restricted subgradient so that the restrictions imposed on $f$ are as minimal as we need. We acknowledge that the existence of restricted subgradients implies convexity in sparse directions, but it does not imply convexity everywhere.
\end{rem}
\begin{rem}
Obviously, if the function $f$ is convex everywhere, then any subgradient of $f$ determines a restricted subgradient of $f$ as well. In general one may need to invoke the axiom of choice to define the restricted subgradient.
\end{rem}  
\begin{rem}
We drop the sparsity level from the notation as it can be understood from the context.
\end{rem}

 With a slight abuse of terminology we call
\begin{align*}
\Breg{f}{\vc{x}'}{\vc{x}} & = f\left(\vc{x}'\right)-f\left(\vc{x}\right) -\left\langle\nabla_{f}\left(\vc{x}\right),\vc{x}'-\vc{x}\right\rangle,
\end{align*} the restricted Bregman divergence of $f:\mathbb{R}^p\mapsto\mathbb{R}$ between points $\vc{x}$ and $\vc{x} '$ where $\nabla_{f}\left(\cdot\right)$ gives a restricted subgradient of $f\left(\cdot\right)$. 

\begin{dfn}[Stable Restricted Linearization]\label{def:D2}
Let $\vc{x}$ be a $k$-sparse vector in $\mathbb{R}^{p}$. For function $f:\mathbb{R}^{p}\mapsto\mathbb{R}$
we \SBedit{define the functions}
\begin{align*}
\alpha_{k}\left(\vc{x}\right) & =\sup\left\{ \frac{1}{\norm{\bsym{\Delta}} _{2}^{2}}\Breg f{\vc{x}+\bsym{\Delta}}{\vc{x}}\,\mid\,\bsym{\Delta}\neq0\mathrm{\ and\ }\left\vert\supp\left(\vc{x}\right) \cup\supp\left(\bsym{\Delta}\right)\right\vert \leq k\right\} 
\intertext{and}\beta_{k}\left(\vc{x}\right) & =\inf\left\{ \frac{1}{\norm{ \bsym{\Delta}} _{2}^{2}}\Breg f{\vc{x}+\bsym{\Delta}}{\vc{x}}\,\mid\,\bsym{\Delta}\neq0\mathrm{\ and\ }\left\vert\supp\left(\vc{x}\right) \cup\supp\left(\bsym{\Delta}\right)\right\vert\leq k\right\}.
\end{align*}
 Then $f\left(\cdot\right)$ is said to have a Stable Restricted Linearization with constant $\mu_k$, or $\mu_k$-\NSCond{}, if $\frac{\alpha_k\left(\vc{x}\right)}{\beta_k\left(\vc{x}\right)}\leq \mu_k$ for all $k$-sparse vectors $\vc{x}$.
\end{dfn}
\SBedit{
\begin{rem}
The SRH and SRL conditions are similar to various forms of the Restricted Strong Convexity (RSC) and Restricted Strong Smoothness (RSS) conditions \citep{Negahban-M-estimators,AgarwalPGD,NLCS_Blumensath,jalali_learning_2011,zhang_sparse_2011} in the sense that they all bound the curvature of the objective function over a restricted set. The SRL condition quantifies the curvature in terms of a (restricted) Bregman divergence similar to RSC and RSS. The quadratic form used in SRH can also be converted to the Bregman divergence form used in RSC and RSS and vice-versa using the mean-value theorem. However, compared to various forms of RSC and RSS conditions SRH and SRL have some important distinctions. The main difference is that the bounds in SRH and SRL conditions are not global constants; only their ratio is required to be bounded globally. Furthermore, unlike the SRH and SRL conditions the variants of RSC and RSS, that are used in convex relaxation methods, are required to hold over a set which is strictly larger than the set of canonical $k$-sparse vectors.

There is also a subtle but important difference regarding the points where the curvature is evaluated at. Since \citet{Negahban-M-estimators} analyze a convex program, rather than an iterative algorithm, they only needed to invoke the RSC and RSS at a neighborhood of the true parameter. In contrast, the other variants of RSC and RSS (see e.g.,\citep{AgarwalPGD,jalali_learning_2011}), as well as our SRH and SRL conditions, require the curvature bounds to hold uniformly over a larger set of points, thereby they are more stringent.
\end{rem}
}
\subsection{Main Theorems}
Now we can state our main results regarding approximation of 
\begin{align}
\vc{x}^{\star}=\arg\min\ f(\vc{x})\text{ s.t. }\norm{\vc{x}}_{0}\leq s,\label{eq:SparseEstim}\end{align}
using the GraSP algorithm.

\begin{thm}\label{thm:thm1}
Suppose that $f$ is a twice continuously differentiable function that has $\mu_{4s}$-\SCond{} with $\mu_{4s}\leq\frac{1+\sqrt{3}}{2}$.
Furthermore, suppose that for some $\epsilon>0$ we have
$\epsilon\leq B_{4s}\left(\vc{u}\right)$ for all $\vc{u}$. Then
$\widehat{\vc{x}}^{\left(i\right)}$, the estimate at the $i$-th iteration,
satisfies
\begin{align*}\norm{\widehat{\vc{x}}^{\left( i\right)}-\vc{x}^{\star}}_{2}\leq2^{-i}\norm{\vc{x}^{\star}}_{2}+\frac{6+2\sqrt{3}}{\epsilon}\norm{\left.\nabla
f\left(\vc{x}^{\star}\right)\right\vert_{\mathcal{I}}}_{2},
\end{align*}
where $\mathcal{I}$ is
the position of the $3s$ largest entries of $\nabla
f\left(\vc{x}^{\star}\right)$ in magnitude.
\end{thm}

\begin{rem} Note that this result indicates that $\nabla f\left(\vc{x}^\star\right)$
determines how accurate the estimate can be. In particular, if the
sparse minimum $\vc{x}^\star$ is sufficiently close to an unconstrained
minimum of $f$ then the estimation error floor is negligible because
$\nabla f\left(\vc{x}^\star\right)$ has small magnitude. This result is
analogous to accuracy guarantees for estimation from noisy measurements in CS~\citep{CandesRombergTao06,NeedellTropp09}.
\end{rem}
\begin{rem}As the derivations required to prove Theorem \ref{thm:thm1} show, the provided accuracy guarantee holds for any $s$-sparse $\vc{x}^\star$, even if it does not obey \eqref{eq:SparseEstim}. Obviously, for arbitrary choices of  $\vc{x}^\star$, $\left.\nabla f\left(\vc{x}^\star\right)\right\vert_\st{I}$ may have a large norm that cannot be bounded properly which implies large errors. In statistical estimation problems, often the true parameter that describes the data is chosen as the target parameter $\vc{x}^\star$ rather than the minimizer of the average loss function as in \eqref{eq:SparseEstim}. In these problems, the approximation error $\norm{\left.\nabla f\left(\vc{x}^\star\right)\right\vert_\st{I}}_2$ has statistical interpretation and can determine the statistical precision of the problem. This property is easy to verify in linear regression problems. We will also show this for the logistic loss as an example in \S\ref{sec:l2-log}.
\end{rem}

Nonsmooth cost functions should be treated differently, since we do not have the luxury of working with Hessian matrices for these type of functions. The following theorem provides guarantees that are similar to those of Theorem \ref{thm:thm1} for nonsmooth cost functions that satisfy the \NSCond{} condition.

\begin{thm}\label{thm:thm2}
Suppose that $f$ is a function that is not necessarily smooth, but it satisfies $\mu_{4s}$-\NSCond{} with $\mu_{4s}\leq\frac{3+\sqrt{3}}{4}$.
Furthermore, suppose that for $\beta_{4s}\left(\cdot\right)$ in Definition \ref{def:D2} there exists some $\epsilon>0$ such that $\beta_{4s}\left(\vc{x}\right)\geq \epsilon$ holds for all $4s$-sparse vectors $\vc{x}$. Then
$\widehat{\vc{x}}^{\left(i\right)}$, the estimate at the $i$-th iteration,
satisfies
\begin{align*}
\norm{\widehat{\vc{x}}^{\left( i\right)}-\vc{x}^{\star}}_{2}\leq2^{-i}\norm{\vc{x}^{\star}}_{2}+\frac{6+2\sqrt{3}}{\epsilon}\norm{\left.\nabla_
f\left(\vc{x}^{\star}\right)\right\vert_{\mathcal{I}}}_{2},
\end{align*}
where $\mathcal{I}$ is
the position of the $3s$ largest entries of $\nabla_
f\left(\vc{x}^{\star}\right)$ in magnitude.
\end{thm}
\begin{rem}
Should the SRH or SRL conditions hold for the objective function, it
is straightforward to convert the \emph{point accuracy} guarantees of
Theorems \ref{thm:thm1} and \ref{thm:thm2}, into accuracy guarantees
in terms of the objective value. First we can use SRH or SRL to bound the Bregman divergence, or its restricted version defined above, for points $\widehat{\vc{x}}^{\left(i\right)}$ and $\vc{x}^\star$. Then we can obtain a bound for the accuracy of the objective value by invoking the results of the theorems. This indirect approach, however, might not lead to sharp bounds and thus we do not pursue the detailed analysis in this work.
\end{rem}

\section{Example: Sparse Minimization of \texorpdfstring{$\ell_2$}{L2}-regularized Logistic Regression}
\label{sec:l2-log}
One of the models widely used in machine learning and
statistics is the logistic model. In this model the relation between the data, represented by a random vector $\vc{a}\in\mathbb{R}^p$, and its associated label, represented by a random binary variable $y\in\left\lbrace 0,1 \right\rbrace$,  is determined by the conditional probability
\begin{align}
\Pr\left\lbrace y \mid \vc{a};\vc{x}\right\rbrace=\frac{\exp\left( y\left\langle\vc{a},\vc{x}\right\rangle\right)}{1+\exp\left(\left\langle\vc{a},\vc{x}\right\rangle\right)},\label{eq:LogisticModel}
\end{align}
where $\vc{x}$ denotes a parameter vector. Then, for a set of $n$ independently drawn data samples $\left\lbrace\left(\vc{a}_i,y_i\right)\right\rbrace_{i=1}^n$ the joint likelihood can be written as a function of $\vc{x}$. To find the maximum likelihood estimate one should maximize this likelihood function, or equivalently minimize the negative log-likelihood, the logistic loss,
\begin{align}
  \label{eq:loglogistic}
  g(\vc{x}) = \frac{1}{n}\sum_{i=1}^{n}\log\left(1+\exp\left(\left\langle\vc{a}_i,\vc{x}\right\rangle\right)\right)-y_i\left\langle\vc{a}_i,\vc{x}\right\rangle.
\end{align} 
It is well-known that $g\left(\cdot\right)$ is strictly convex for
$p\leq n$ provided that the associated design matrix,
$\mx{A} = \left[\vc{a}_1\:\vc{a}_2\:\ldots\:\vc{a}_n\right]^\mathrm{T}$,
is full-rank. However, in many important applications (e.g., feature
selection) the problem can be underdetermined (i.e., $n<p$). In these
scenarios the logistic loss is merely convex and it does not have a
unique minimum. Furthermore, it is possible, especially in
underdetermined problems, that the observed data is linearly
separable. In that case one can achieve arbitrarily small loss values
by tending the parameters to infinity along certain directions. To
compensate for these drawbacks the logistic loss is usually
regularized by some penalty term~\citep{hastie2009elements,Bunea08}.

One of the candidates for the penalty function is the (squared)
$\ell_2$-norm of $\vc{x}$ (i.e., $\norm{\vc{x}}_{2}^{2}$). Considering a
positive penalty coefficient $\eta$ the regularized loss is
\begin{align*}
f\left(\vc{x}\right)&=g(\vc{x})+\frac{\eta}{2}\norm{\vc{x}}_{2}^{2}.
\end{align*}
 For any convex
$g\left(\cdot\right)$ this regularized loss is guaranteed to be
$\eta$-strongly convex, thus it has a unique minimum. Furthermore, the
penalty term implicitly bounds the length of the minimizer thereby
resolving the aforementioned problems. Nevertheless, the $\ell_2$
penalty does not promote sparse solutions. Therefore, it is often
desirable to impose an explicit sparsity constraint, in addition to
the $\ell_2$ regularizer.

\subsection{Verifying \SCond{} for \texorpdfstring{$\ell_2$}{L2}-regularized logistic loss}
It is easy to show that the Hessian of the logistic loss at any point
$\vc{x}$ is given by $\mx{H}_{g}\! \left( \vc{x} \right) = \frac{1}{4n}
\mx{A}^\mathrm{T}\bsym{\Lambda}\mx{A}$, where $\bsym{\Lambda}$ is an $n\times n$
diagonal matrix whose diagonal entries
$\Lambda_{ii}=\text{sech}^2\frac{1}{2}\left\langle\vc{a}_i,\vc{x}\right\rangle$ with $\text{sech}\left(\cdot\right)$ denoting the \emph{hyperbolic secant} function. Note
that
$\mathbf{0}\preccurlyeq\mx{H}_{g}\left(\vc{x}\right)\preccurlyeq\frac{1}{4n}\mx{A}^\mathrm{T}\mx{A}$. Therefore,
if $\mx{H}_{\eta}\left(\vc{x}\right)$ denotes the Hessian of the $\ell_2$-regularized
logistic loss, we have
\begin{align}
  \forall \vc{x} ,\bsym{\Delta} & & \eta\norm{\bsym{\Delta}}_{2}^{2}\leq\bsym{\Delta}^\mathrm{T}\mx{H}_{\eta}\left(\vc{x}\right)\bsym{\Delta}\leq\frac{1}{4n}\norm{\mx{A}\bsym{\Delta}}_{2}^{2}+\eta\norm{\bsym{\Delta}}_{2}^{2}. \label{eq:HessEigs}
\end{align}
To verify \SCond{}, the upper and lower bounds achieved at $k$-sparse
vectors $\bsym{\Delta}$ are of particular interest. It only remains to find an
appropriate upper bound for $\norm{\mx{A}\bsym{\Delta}}_{2}^{2}$ in terms
of $\norm{\bsym{\Delta}}_{2}^{2}$. To this end we use the following result on
Chernoff bounds for random matrices due to \citet{UserFriendlyTails}.

\begin{thm}[Matrix Chernoff \citep{UserFriendlyTails}]\label{thm:Tropp}
	Consider a finite sequence $\lbrace\mx{M}_i\rbrace$ of $k\times k$, independent, random, self-adjoint matrices that satisfy
	\begin{align*}
		\mx{M}_i\succcurlyeq\mathbf{0} \quad and \quad \lambda_\mathrm{max}\left(\mx{M}_i\right)\leq R \quad almost \: surely.
	\end{align*} Let $\theta_\mathrm{max}:=\lambda_\mathrm{max}\left(\sum_i\mathbb{E}\left[\mx{M}_i\right]\right)$. Then for $\tau\geq 0$,
	\begin{align*}
		\Pr \left\lbrace \lambda_\mathrm{max} \left( \sum _{i}\mx{M}_i \right) \geq \left(1+\tau\right)\theta_\mathrm{max} \right\rbrace \leq & k\exp\left(\frac{\theta_\mathrm{max}}{R}\left(\tau-(1+\tau\right)\log\left(1+\tau\right) \right).
	\end{align*}		
\end{thm}

As stated before, in a standard logistic model data samples $\left\lbrace\vc{a}_i\right\rbrace$ are supposed to be independent instances of a random vector $\vc{a}$. In order to apply Theorem \ref{thm:Tropp} we need to make the following extra assumptions:
\begin{asmp*}
For every $\st{J}\subseteq\left[ p\right]$ with $\left\vert\st{J}\right\vert= k$,
	\begin{enumerate}[label={(\roman*)}]
	\item \label{asmp:BoundedOperator} we have	$\norm{\left.\vc{a}\right\vert_\st{J}}_{2}^{2}\leq R$ almost surely, and
	\item \label{asmp:NonDegenerate} none of the matrices $\mx{P}^\mathrm{T}_{\st{J}}\mathbb{E}\left[\vc{a}\vc{a}^\mathrm{T}\right]\mx{P}_{\st{J}}$ is the zero matrix.
	\end{enumerate}
\end{asmp*}
\SBedit{We define $\theta_{\max}^\st{J}:=\lambda_{\max}\left(\mx{P}^\mathrm{T}_\st{J}\mx{C}\mx{P}_\st{J}\right)$, where $\mx{C}=\mathbb{E}\left[\vc{a}\vc{a}^\mathrm{T}\right]$, and let 
\begin{alignat*}{3}
	\overline{\theta} & :=\underset{\st{J} \subseteq [p] \:,\: \left\vert\st{J}\right\vert=k
  }{\max} \theta_{\max}^\st{J}&
 \quad\text{ and }\quad & \widetilde{\theta}\ &:=\underset{\st{J} \subseteq [p] \:,\: \left\vert\st{J}\right\vert=k}{\min} \theta_{\max}^\st{J}.
\end{alignat*}
To simplify the notation henceforth we let $h\left(\tau\right)=\left(1+\tau\right)\log\left(1+\tau\right)-\tau$.
\begin{cor}\label{cor:vrfySRH}
With the above assumptions, if $n \geq
R\left( \log k + k\left( 1+\log
\frac{p}{k}\right) - \log \varepsilon \right)/\left(\widetilde{\theta} h\left(\tau\right)\right)$ for some $\tau>0$ and $\varepsilon\in\left(0,1\right)$, then with probability at least $1-\varepsilon$ the $\ell_2$-regularized logistic loss has $\mu_k$-SRH with $\mu_k\leq 1+\frac{1+\tau}{4\eta}\overline{\theta}$.
\end{cor}
\begin{proof}
For any set of $k$ indices $\st{J}$ let $\mx{M}^\st{J}_i=\left.\vc{a}_i\right\vert_\st{J}\left.\vc{a}_{i}\right\vert_\st{J}^\mathrm{T}=\mx{P}^\mathrm{T}_\st{J}\vc{a}_i\vc{a}_{i}^\mathrm{T}\mx{P}_\st{J}$.
The independence of the vectors $\vc{a}_i$ implies that the
matrix
\begin{alignat*}{1}
\mx{A}^\mathrm{T}_{\st{J}}\mx{A}_{\st{J}} &= \sum_{i=1}^{n} \left.\vc{a}_{i}\right\vert_{\st{J}}
\left.\vc{a}_{i}\right\vert_{\st{J}}^\mathrm{T}\\
 & =\sum_{i=1}^n \mx{M}^\st{J}_i
\end{alignat*}
is a sum of $n$ independent, random,
self-adjoint matrices. Assumption \ref{asmp:BoundedOperator} implies that
$\lambda_\mathrm{max}\left(\mx{M}_i^\st{J}\right)=\norm{\left.\vc{a}_{i}\right\vert_\st{J}}_2^2\leq
R$ almost surely. Furthermore, we have 
\begin{alignat*}{1}
\lambda_{\max}\left(\sum_{i=1}^n \mathbb{E}\left[\mx{M}_i^\st{J}\right]\right) &= \lambda_{\max}\left(\sum_{i=1}^n \mathbb{E}\left[\mx{P}^\mathrm{T}_\st{J}\vc{a}_i\vc{a}_i^\mathrm{T}\mx{P}_\st{J}\right]\right)\\ &= \lambda_{\max}\left(\sum_{i=1}^n \mx{P}^\mathrm{T}_\st{J}\mathbb{E}\left[\vc{a}_i\vc{a}_i^\mathrm{T}\right]\mx{P}_\st{J}\right)\\
&= \lambda_{\max}\left(\sum_{i=1}^n \mx{P}^\mathrm{T}_\st{J}\mx{C}\mx{P}_\st{J}\right)\\
&= n\lambda_{\max}\left(\mx{P}^\mathrm{T}_\st{J}\mx{C}\mx{P}_\st{J}\right)\\
&= n\theta_{\max}^\st{J}.
\end{alignat*}
Hence, for any fixed index set $\st{J}$ with $\left\vert\st{J}\right\vert = k$ we may apply Theorem \ref{thm:Tropp} for
$\mx{M}_i=\mx{M}_i^\st{J}$, $\theta_{\max}=n\theta_{\max}^\st{J}$, and $\tau > 0$ to obtain 
\begin{alignat*}{1}
		\Pr \left\lbrace \lambda_\mathrm{max} \left( \sum
                _{i=1}^{n}\mx{M}_i^\st{J} \right) \geq
                \left(1+\tau\right)n\theta_{\max}^\st{J} \right\rbrace \leq &
                k\exp\left(-\frac{n\theta_{\max}^\st{J}h\left(\tau\right)}{R}\right).
\end{alignat*}
Furthermore, we can write
\begin{alignat}{1}
\Pr \left\lbrace \lambda_\mathrm{max} \left(\mx{A}_\st{J}^\mathrm{T}\mx{A}_\st{J}\right) \geq
                \left(1+\tau\right)n\overline{\theta} \right\rbrace &=\Pr \left\lbrace \lambda_\mathrm{max} \left( \sum_{i=1}^{n}\mx{M}_i^\st{J} \right) \geq        \left(1+\tau\right)n\overline{\theta} \right\rbrace \nonumber \\
				 & \leq \Pr \left\lbrace \lambda_\mathrm{max} \left( \sum
                _{i=1}^{n}\mx{M}_i^\st{J} \right) \geq
                \left(1+\tau\right)n\theta_{\max}^\st{J} \right\rbrace \nonumber \\ 
                & \leq k\exp\left(-\frac{n\theta_{\max}^\st{J}h\left(\tau\right)}{R}\right) \nonumber \\
                & \leq k\exp\left(-\frac{n\widetilde{\theta}h\left(\tau\right)}{R}\right). \label{eq:SingleSetConcentration}	
\end{alignat}
Note that Assumption \ref{asmp:NonDegenerate} guarantees that $\widetilde{\theta}>0$, and thus the above probability bound will not be vacuous for sufficiently large $n$. To ensure a uniform guarantee for all $\tbinom{p}{k}$ possible choices of $\st{J}$ we can
use the union bound to obtain
\begin{align*} 
  \Pr \left\lbrace \bigvee_{\substack{\st{J} \subseteq \left[p\right]\\ \left\vert\st{J}
    \right\vert=k}}\lambda_\mathrm{max}\!
  \left(\mx{A}^\mathrm{T}_{\st{J}}\mx{A}_{\st{J}}\right) \! \geq \!
  \left(1\!+\!\tau\right)n\overline{\theta} \! \right\rbrace &\leq \sum_{\substack{\st{J} \subseteq \left[p\right]\\ \left\vert\st{J}\right\vert=k}} \Pr \left\lbrace \lambda_\mathrm{max}\!
  \left(\mx{A}^\mathrm{T}_{\st{J}}\mx{A}_{\st{J}}\right) \! \geq \!
  \left(1\!+\!\tau\right)n\overline{\theta} \! \right\rbrace \\
  &\leq k \binom{p}{k}
  \exp\left(-\frac{n\widetilde{\theta}h\left(\tau\right)}{R}\right)
  \\ &\leq k \left(\frac{pe}{k}\right)^k
  \exp\left(-\frac{n\widetilde{\theta}h\left(\tau\right)}{R}\right)
  \\  &= \exp\left(\!\log k \!+\! k\!+\! k\log\frac{p}{k}\! -
  \!\frac{n\widetilde{\theta}h\left(\tau\right)}{R}\!\right). 
\end{align*}
Therefore, for $\varepsilon \in \left( 0,1 \right)$ and $n \geq
R\left( \log k + k\left( 1+\log
\frac{p}{k}\right) - \log \varepsilon \right)/\left(\widetilde{\theta} h\left(\tau\right)\right)$ it follows from \eqref{eq:HessEigs} that for any $\vc{x}$
and any $k$-sparse $\bsym{\Delta}$,
\begin{align*}	\eta\norm{\bsym{\Delta}}_2^2\leq\bsym{\Delta}^\mathrm{T}\mx{H}_\eta\left(\vc{x}\right)\bsym{\Delta}\leq\left(\eta+\frac{1+\tau}{4}\overline{\theta}\right)\norm{\bsym{\Delta}}_2^2
\end{align*}
	holds with probability at least $1-\varepsilon$. Thus, the $\ell_2$-regularized logistic loss has an \SCond{} constant $\mu_k\leq 1+\frac{1+\tau}{4\eta}\overline{\theta}$ with probability $1-\varepsilon$.
\end{proof}
\begin{rem}One implication of this result is that for a regime in
which $k$ and $p$ grow sufficiently large while $\frac{p}{k}$ remains
constant one can achieve small failure rates provided that
$n=\varOmega \left(Rk\log \frac{p}{k}\right)$. Note that $R$ is deliberately included in the argument of the order function because in general $R$ depends on $k$. In other words, the above analysis may require $n=\Omega\left(k^2\log\frac{p}{k}\right)$ as the sufficient number of observations. This bound is a consequence of using Theorem \ref{thm:Tropp}, but to the best of our knowledge, other results regarding the extreme eigenvalues of the average of independent random PSD matrices also yield an $n$ of the same order. If matrix $\mx{A}$ has certain additional properties (e.g., independent and sub-Gaussian entries), however, a better rate of $n=\Omega\left(k\log\frac{p}{k}\right)$ can be achieved without using the techniques mentioned above.
\end{rem}
\begin{rem}
The analysis provided here is not specific to the $\ell_2$-regularized logistic loss and can be readily extended to any other $\ell_2$-regularized GLM loss whose log-partition function has a Lipschitz-continuous derivative.
\end{rem}
}

\subsection{Bounding the approximation error}
We are going to bound $\norm{\left.\nabla f\left(\vc{x}^\star\right)\right\vert_\st{I}}_2$ which controls the approximation error in the statement of Theorem \ref{thm:thm1}. In the case of case of $\ell_2$-regularized logistic loss considered in this section we have
\begin{align*}
\nabla f\left(\vc{x}\right)&=\sum_{i=1}^n \left(\frac{1}{1+\exp\left(-\left\langle\vc{a}_i,\vc{x}\right\rangle\right)}-y_i\right)\vc{a}_i+ \eta \vc{x}.
\end{align*}
Denoting $\tfrac{1}{1+\exp\left(-\left\langle \vc{a}_i,\vc{x}^\star\right\rangle\right)}-y_i$ by $v_i$ for $i=1,2,\ldots,n$ then we can deduce
\begin{align*}
\norm{\left.\nabla f\left(\vc{x}^\star\right)\right\vert_\st{I}}_2 &= \norm{\frac{1}{n}\sum_{i=1}^n v_i \left.\vc{a}_i\right\vert_\st{I}+\eta\left.\vc{x}^\star\right\vert_\st{I}}_2 \\
&=\norm{\frac{1}{n}\mx{A}^\mathrm{T}_\st{I}\vc{v}+\eta \left.\vc{x}^\star\right\vert_\st{I}}_2\\
&\leq \frac{1}{n}\norm{\mx{A}^\mathrm{T}_\st{I}}\norm{\vc{v}}_2+\eta\norm{\left.\vc{x}^\star\right\vert_\st{I}}_2\\
 &\leq \frac{1}{\sqrt{n}}\norm{\mx{A}_\st{I}}\sqrt{\frac{1}{n}\sum_{i=1}^nv_i^2}+\eta\norm{\left.\vc{x}^\star\right\vert_\st{I}}_2,
\end{align*}
where $\vc{v} = \left[v_1\,v_2\, \ldots v_n\right]^\mathrm{T}$. Note that $v_i$'s are $n$ independent copies of the random variable $v = \frac{1}{1+\exp\left(-\left\langle \vc{a},\vc{x}^\star\right\rangle\right)}-y$ that is zero-mean and always lie in the interval $\left[-1,1\right]$. Therefore, applying the Hoeffding's inequality yields
\begin{align*}
	\Pr\left\lbrace\frac{1}{n}\sum_{i=1}^n v_i^2 \geq \left(1+c\right)\sigma^2_v \right\rbrace &\leq \exp\left(-2nc^2\sigma^4_v\right),
\end{align*}
where $\sigma^2_v=\mathbb{E}\left[v^2\right]$ is the variance of $v$. Furthermore, using the logistic model \eqref{eq:LogisticModel} we can deduce
\begin{align*}
	\sigma^2_v &= \mathbb{E}\left[v^2\right]\\
	&=\mathbb{E}\left[\mathbb{E}\left[v^2\mid\vc{a}\right]\right]\\
	&=\mathbb{E}\left[\mathbb{E}\left[\left(y-\mathbb{E}\left[y\mid \vc{a}\right]\right)^2\mid\vc{a}\right]\right]\\
	&=\mathbb{E}\left[\mathrm{var}\left(y \mid \vc{a}\right)\right]\\
	&=\mathbb{E}\left[\frac{1}{1+\exp\left(\left\langle\vc{a},\vc{x}^\star\right\rangle\right)}\times\frac{\exp\left(\left\langle\vc{a},\vc{x}^\star\right\rangle\right)}{1+\exp\left(\left\langle\vc{a},\vc{x}^\star\right\rangle\right)}\right]&\text{(because } y\mid\vc{a}\sim\text{Bernoulli as in \eqref{eq:LogisticModel})}\\
	&=\mathbb{E}\left[\frac{1}{2+\exp\left(\left\langle\vc{a},\vc{x}^\star\right\rangle\right)+\exp\left(-\left\langle\vc{a},\vc{x}^\star\right\rangle\right)} \right]\\
&\leq\frac{1}{4}&\text{(because } \exp\left(t\right)+\exp\left(-t\right)\geq 2\text{)}.
\end{align*}
Therefore, we have $\frac{1}{n}\sum_{i=1}^n v_i^2 <\frac{1}{4}$ with high probability. As in the previous subsection one can also bound $\frac{1}{\sqrt{n}}\norm{\mx{A}_\st{I}}=\sqrt{\frac{1}{n}\lambda_{\max}\left(\mx{A}^\mathrm{T}_\st{I}\mx{A}_\st{I}\right)}$ using \eqref{eq:SingleSetConcentration} with $k=\left\vert\st{I}\right\vert=3s$. Hence, with high probability we have
\begin{align*}
	\norm{\left.\nabla f\left(\vc{x^\star}\right)\right\vert_\st{I}}_2&\leq \frac{1}{2}\sqrt{\left(1+\tau\right)\overline{\theta}}+\eta\norm{\vc{x}^\star}_2.
\end{align*}
Interestingly, this analysis can also be extended to the GLMs whose log-partition function $\psi\left(\cdot\right)$ obeys $0\leq\psi''\left(t\right)\leq C$ for all $t$ with $C$ being a positive constant. For these models the approximation error can be bounded in terms of the variance of $v_\psi =\psi'\left(\left\langle\vc{a},\vc{x}^\star\right\rangle\right)-y$.

\section{Experimental Results}
\label{sec:experiments}
\subsection*{Synthetic Data}
Algorithms that are used for sparsity-constrained estimation or
optimization often induce sparsity using different types of
regularizations or constraints. Therefore, the {\em optimized} objective function may vary from one algorithm to
another, even though all of these algorithms try to estimate the same
sparse parameter and sparsely optimize the same original
objective. Because of the discrepancy in the optimized objective
functions it is generally difficult to compare performance of these
algorithms. Applying algorithms on real data generally produces even
less reliable results because of the unmanageable or unknown
characteristics of the real data. Hence, we restrict our simulations to application of the logistic model
on synthetically generated data, and evaluate performance of
GraSP based on the value of the
original objective and the estimation accuracy for a sparse parameter.

In our simulations the sparse parameter of interest $\vc{x}^\star$ is
a $p=1000$ dimensional vector that has $s=10$ nonzero entries drawn
independently from the standard Gaussian distribution. An intercept
$c\in\mathbb{R}$ is also considered which is drawn independently of
the other parameters according to the standard Gaussian
distribution. Each data sample is an independent instance of the
random vector $\vc{a}= \left[a_1,a_2,\ldots,a_p\right]^\mathrm{T}$
generated by an autoregressive process \citep{Hamilton94} determined
by
\begin{align*}
a_{j+1}&=\rho a_j + \sqrt{1-\rho^2}z_j,& \text{for all }j\in\left[ p-1\right]
\end{align*}
with $a_1\sim\mathcal{N}\left(0,1\right)$,
$z_j\sim\mathcal{N}\left(0,1\right)$, and $\rho\in\left[0,1\right]$
being the correlation parameter. The data model we describe and use
above is identical to the experimental model used in
\citep{AgarwalPGD}, except that we adjusted the coefficients to ensure
that $\mathbb{E}\left[a_j^2\right]=1$ for all $j\in\left[ p\right]$.
The data labels, $y\in\left\lbrace 0,1\right\rbrace$ are then drawn
randomly according to the Bernoulli distribution with
\begin{align*}
\Pr\left\{y=0\mid \vc{a}\right\}&=1/\left(1+\exp\left(\left\langle\vc{a},\vc{x}^{\star}\right\rangle+c\right)\right).
\end{align*} 

We compared GraSP to the LASSO algorithm implemented in the GLMnet package~\citep{GLMnet-FHT2009}, as well as the Orthogonal Matching Pursuit method dubbed Logit-OMP~\citep{lozano_group_2011}. To isolate the effect of $\ell_2$-regularization, both LASSO and the basic implementation of GraSP did not consider additional $\ell_2$-regularization terms. To analyze the effect of an additional $\ell_2$-regularization we also evaluated the performance of GraSP with $\ell_2$-regularized logistic loss, as well as the logistic regression with elastic net (i.e., mixed $\ell_1$-$\ell_2$) penalty also available in the GLMnet package. We configured the GLMnet software to produce $s$-sparse solutions for a fair comparison. For the elastic net penalty $\left(1-\omega\right)\norm{\vc{x}}^2_2/2+\omega\norm{\vc{x}}_1$ we considered the ``mixing parameter'' $\omega$ to be 0.8. For the $\ell_2$-regularized logistic loss we considered $\eta=\left(1-\omega\right)\sqrt{\tfrac{\log p}{n}}$. For each choice of the number of measurements $n$ between 50 and 1000 in steps of size 50, and $\rho$ in the set \mbox{$\left\{ 0, \tfrac{1}{3}, \tfrac{1}{2},\tfrac{\sqrt{2}}{2}\right\}$} we generate the data and the associated labels and apply the algorithms. The average performance is measured over 200 trials for each pair of $\left(n,\rho\right)$.

\begin{figure}[t]
	\centering
	\subfloat[$\rho=0$]{\label{fig:rho1}\includegraphics[width=0.5\textwidth]{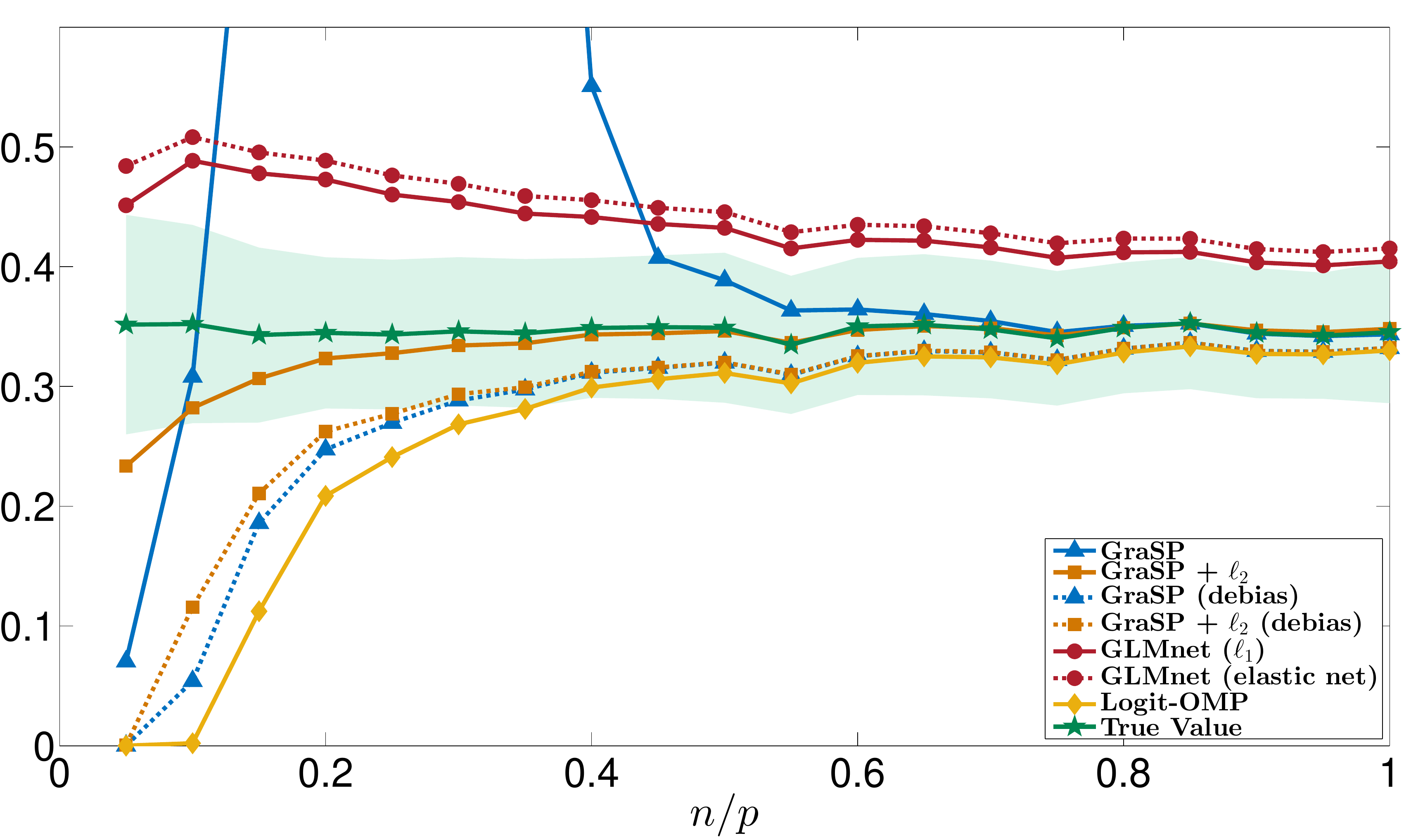}}
	\subfloat[$\rho=\nicefrac{1}{3}$]{\label{fig:rho2}\includegraphics[width=0.5\textwidth]{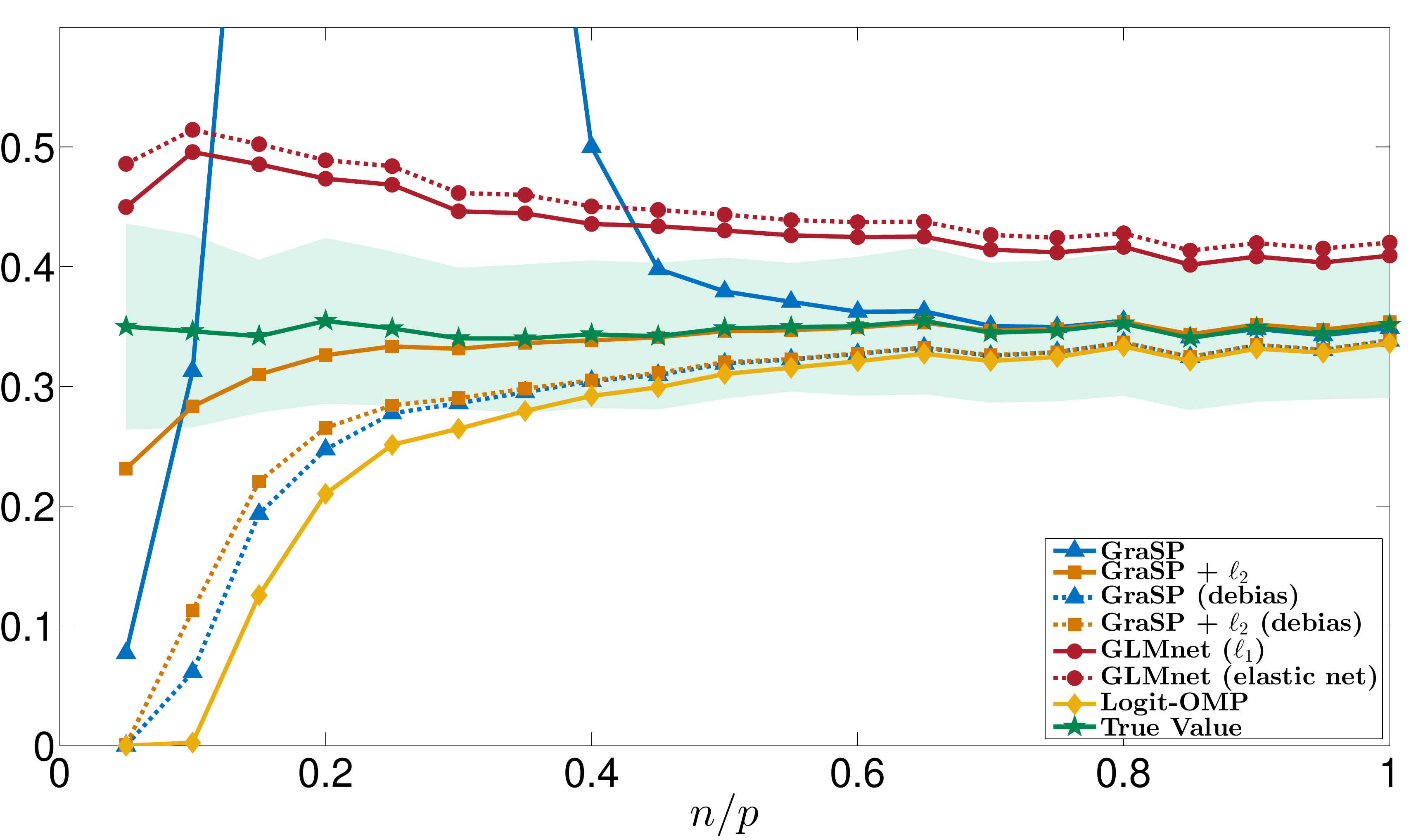}}
	\hfill\subfloat[$\rho=\nicefrac{1}{2}$]{\label{fig:rho3}\includegraphics[width=0.5\textwidth]{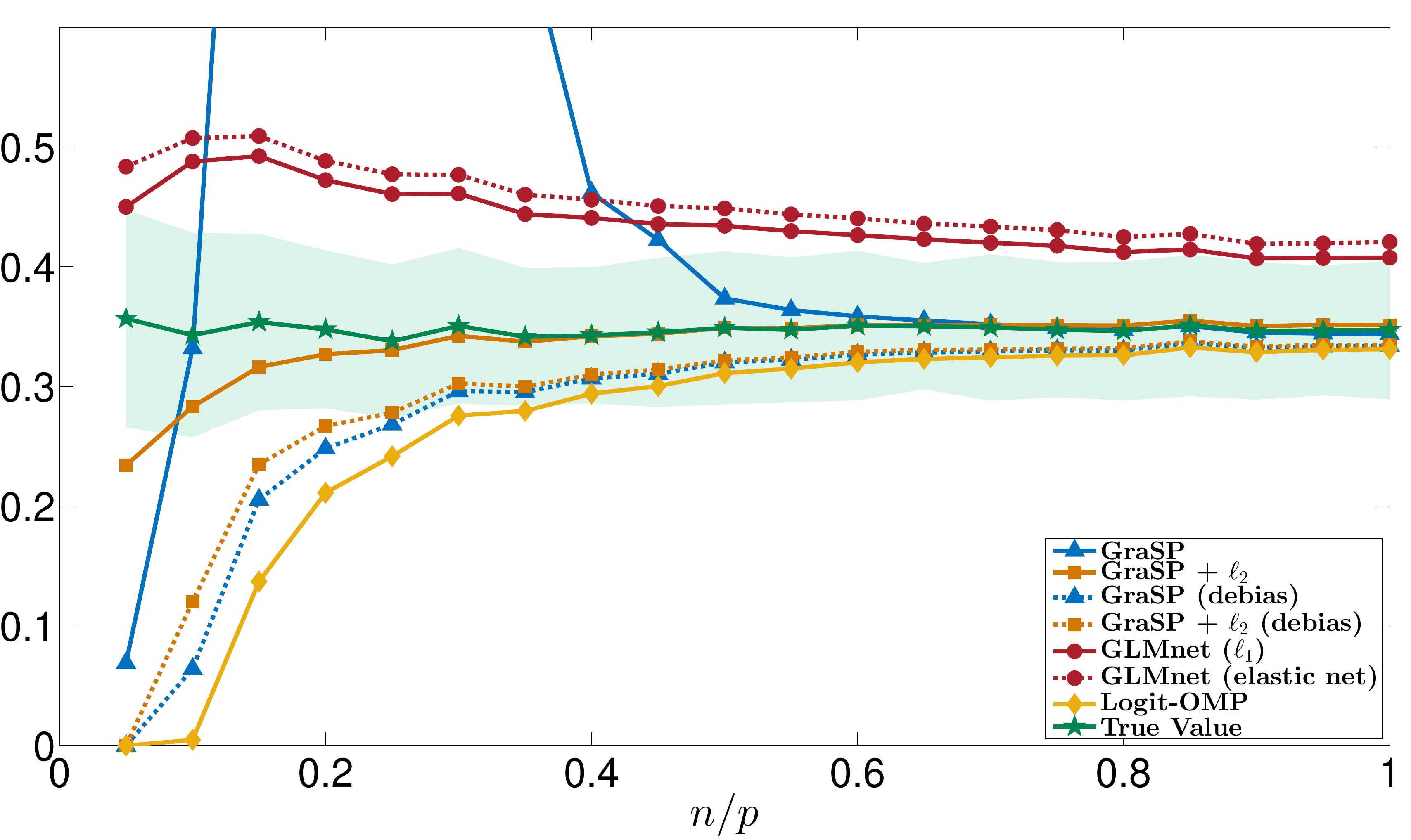}}
	\subfloat[$\rho=\nicefrac{\sqrt{2}}{2}$]{\label{fig:rho4}\includegraphics[width=0.5\textwidth]{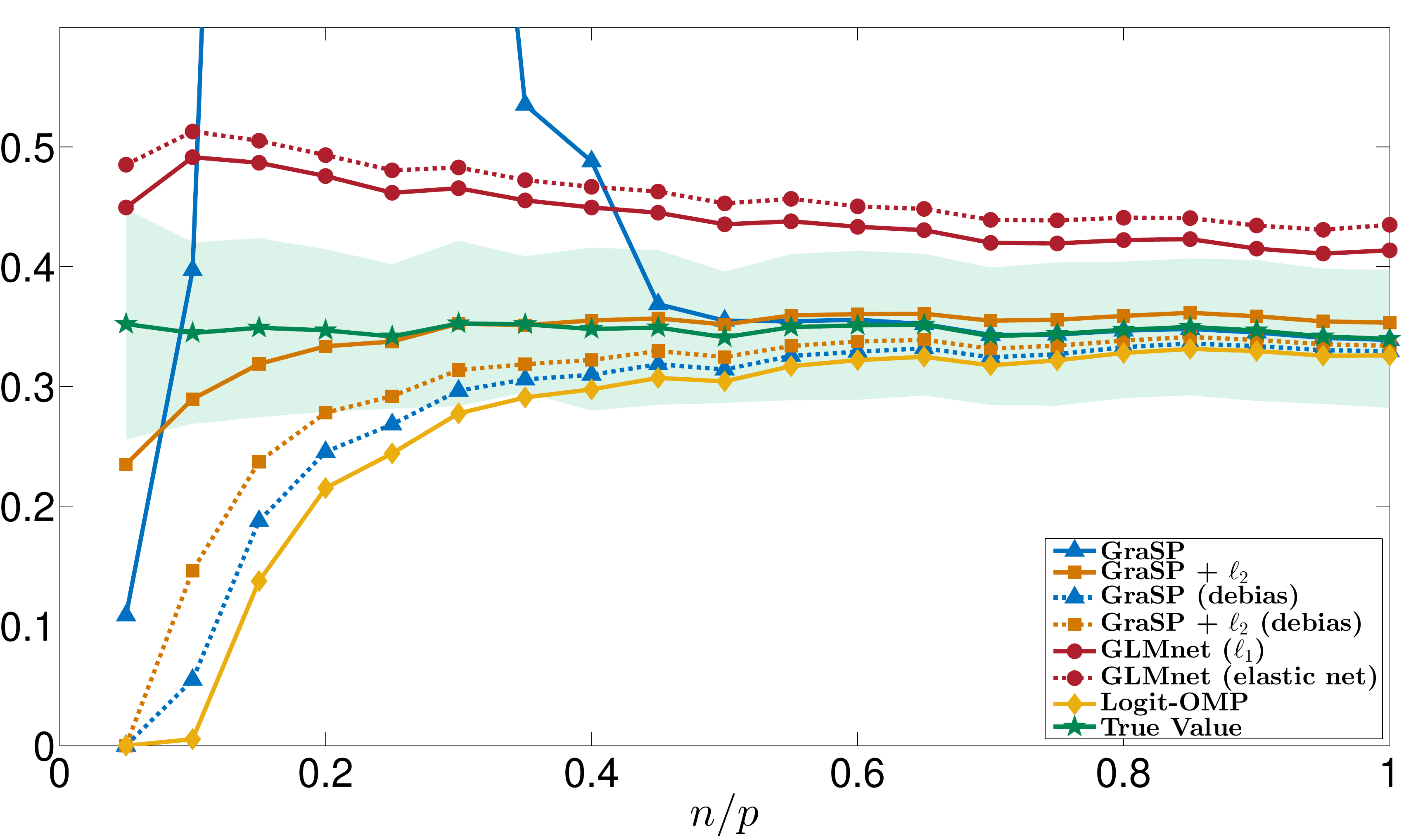}}
	\caption{Comparison of the average (empirical) logistic loss at solutions obtained via GraSP, GraSP with $\ell_2$-penalty, LASSO, the elastic-net regularization, and Logit-OMP. The results of both GraSP methods with ``debiasing'' are also included. The average loss at the true parameter and one standard deviation interval around it are plotted as well.}
	\label{fig:experiments}
\end{figure}

Fig. \ref{fig:experiments} compares the average value of the empirical logistic loss achieved by each of the considered algorithms for a wide range of ``sampling ratio'' $n/p$. For GraSP, the curves labelled by GraSP and GraSP + $\ell_2$ corresponding to the cases where the algorithm is applied to unregularized and $\ell_2$-regularized logistic loss, respectively. Furthermore, the results of GLMnet for the LASSO and the elastic net regularization are labelled by GLMnet ($\ell_1$) and GLMnet (elastic net), respectively. The simulation result of the Logit-OMP algorithm is also included. To contrast the obtained results we also provided the average of empirical logistic loss evaluated at the true parameter and one standard deviation above and below this average on the plots. Furthermore, we evaluated performance of GraSP with the debiasing procedure described in \S\ref{ssec:AlgDescription}.

As can be seen from the figure at lower values of the sampling ratio GraSP is not accurate and does not seem to be converging. This behavior can be explained by the fact that without regularization at low sampling ratios the training data is \emph{linearly separable} or has very few mislabelled samples. In either case, the value of the loss can vary significantly even in small neighborhoods. Therefore, the algorithm can become too sensitive to the pruning step at the end of each iteration. At larger sampling ratios, however, the loss from GraSP begins to decrease rapidly, becoming effectively identical to the loss at the true parameter for $n/p > 0.7$. The results show that unlike GraSP, Logit-OMP performs gracefully at lower sampling ratios. At higher sampling ratios, however, GraSP appears to yield smaller bias in the loss value. Furthermore, the difference between the loss obtained by the LASSO and the loss at the true parameter never drops below a certain threshold, although the convex method exhibits a more stable behaviour at low sampling ratios.

Interestingly, GraSP becomes more stable at low sampling ratios when the logistic loss is regularized with the $\ell_2$-norm. However, this stability comes at the cost of a bias in the loss value at high sampling ratios that is particularly pronounced in Fig. \ref{fig:rho4}. Nevertheless, for all of the tested values of $\rho$, at low sampling ratios GraSP+$\ell_2$ and at high sampling ratios GraSP are consistently closer to the true loss value compared to the other methods. Debiasing the iterates of GraSP also appears to have a stabilizing effect at lower sampling ratios. For GraSP with $\ell_2$ regularized cost, the debiasing particularly reduced the undesirable bias at $\rho=\frac{\sqrt{2}}{2}$.

\begin{figure}[t]
	\centering
	\subfloat[$\rho=0$]{\label{fig:rho1_RRE}\includegraphics[width=0.5\textwidth]{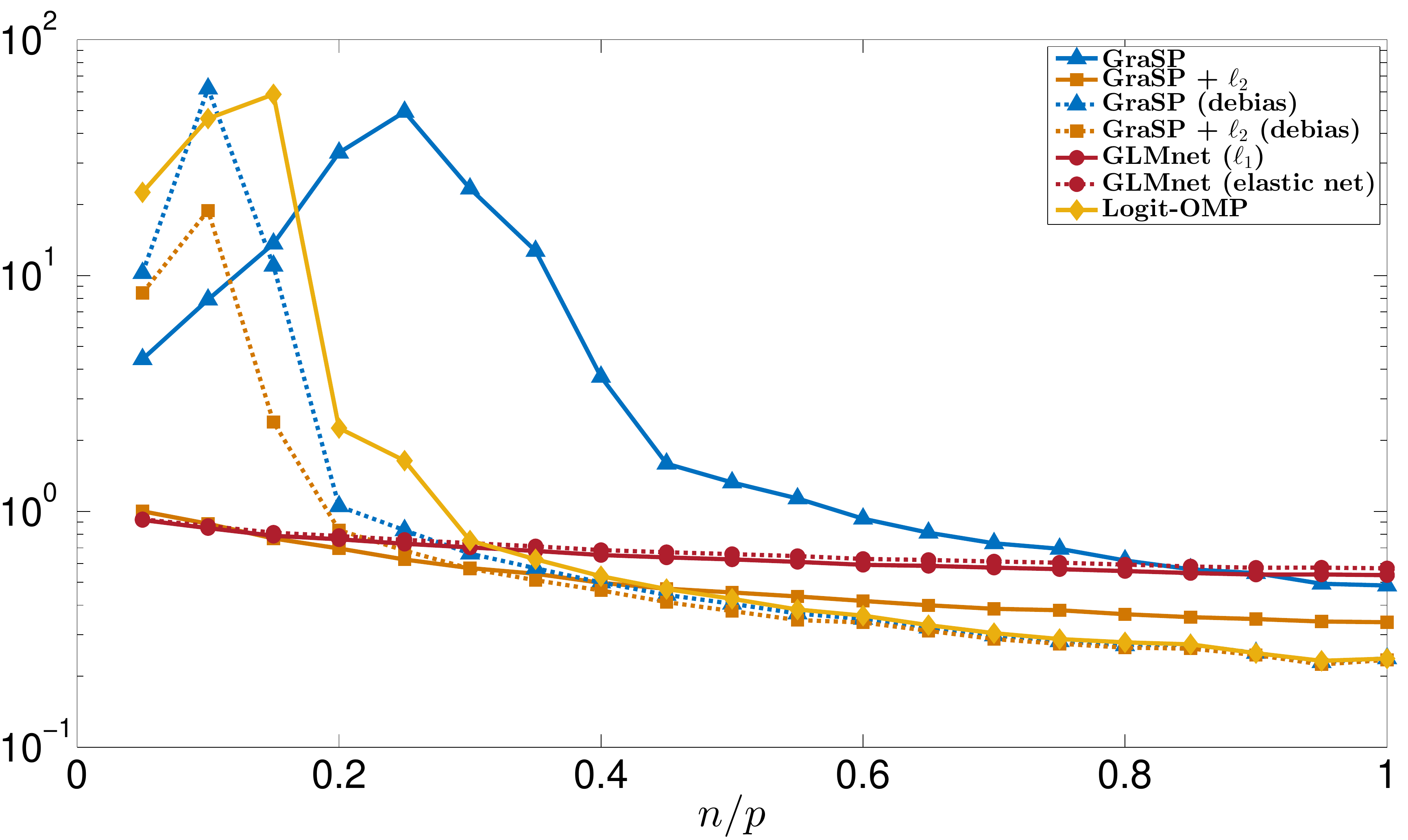}}
	\subfloat[$\rho=\nicefrac{1}{3}$]{\label{fig:rho2_RRE}\includegraphics[width=0.5\textwidth]{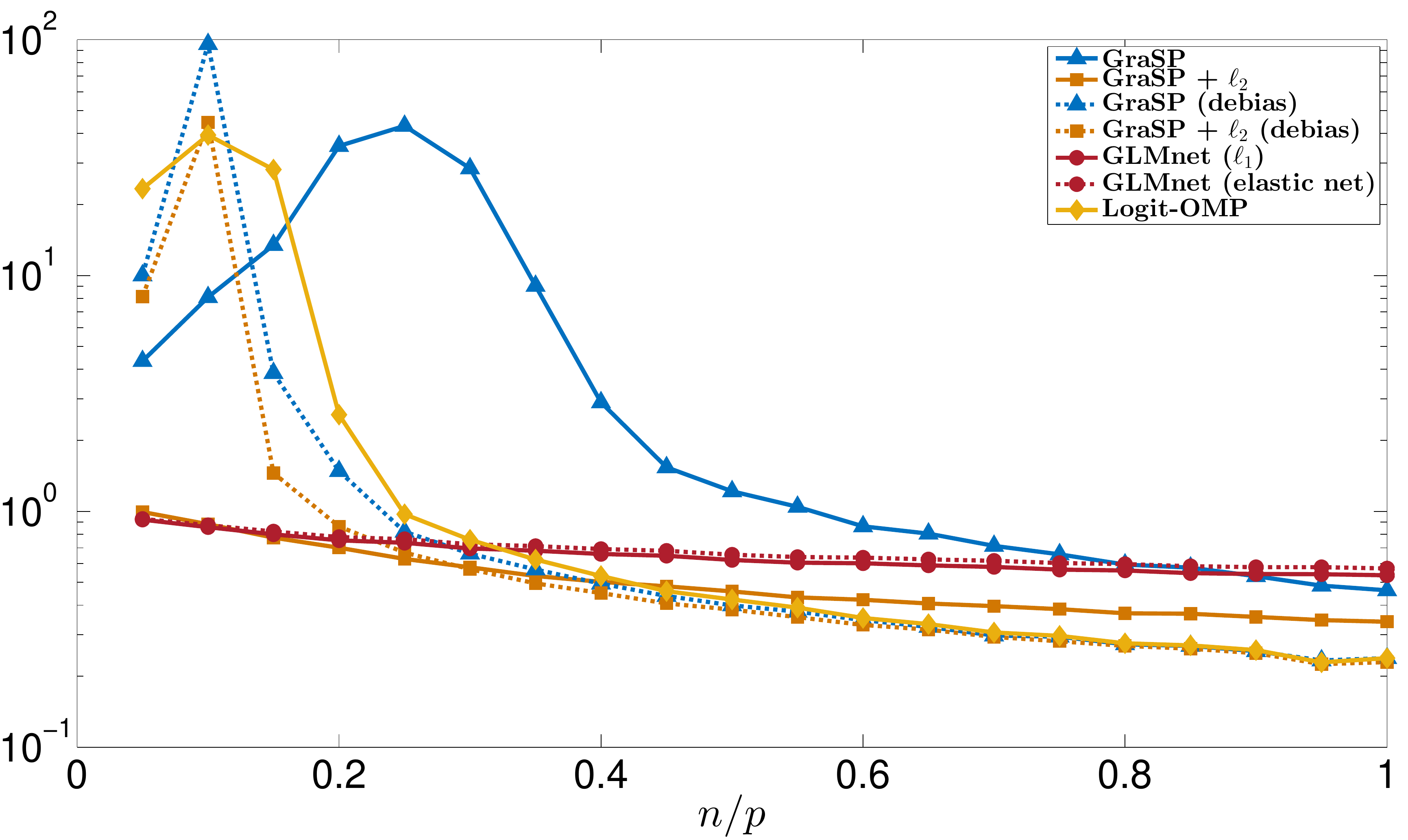}}
	\hfill\subfloat[$\rho=\nicefrac{1}{2}$]{\label{fig:rho3_RRE}\includegraphics[width=0.5\textwidth]{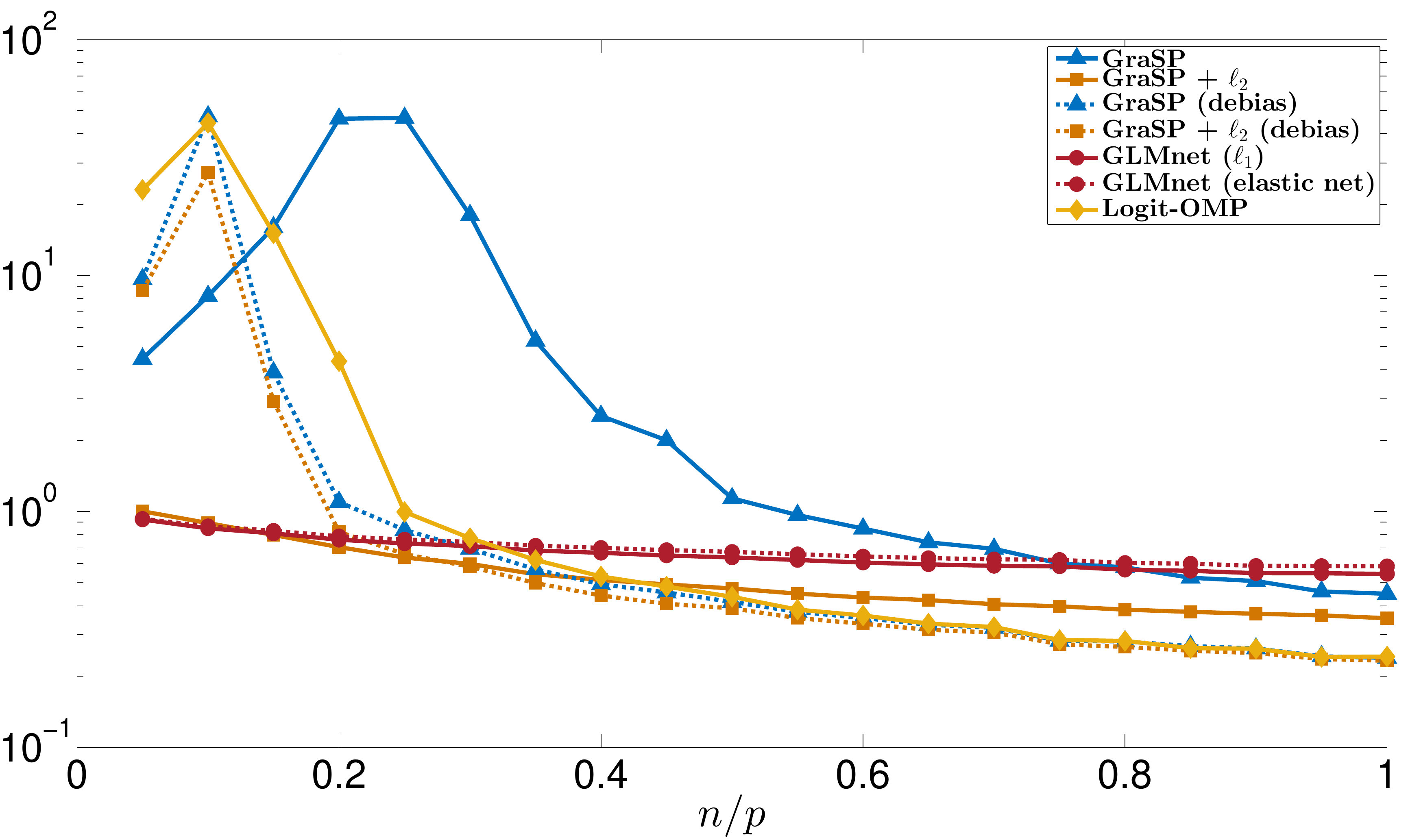}}
	\subfloat[$\rho=\nicefrac{\sqrt{2}}{2}$]{\label{fig:rho4_RRE}\includegraphics[width=0.5\textwidth]{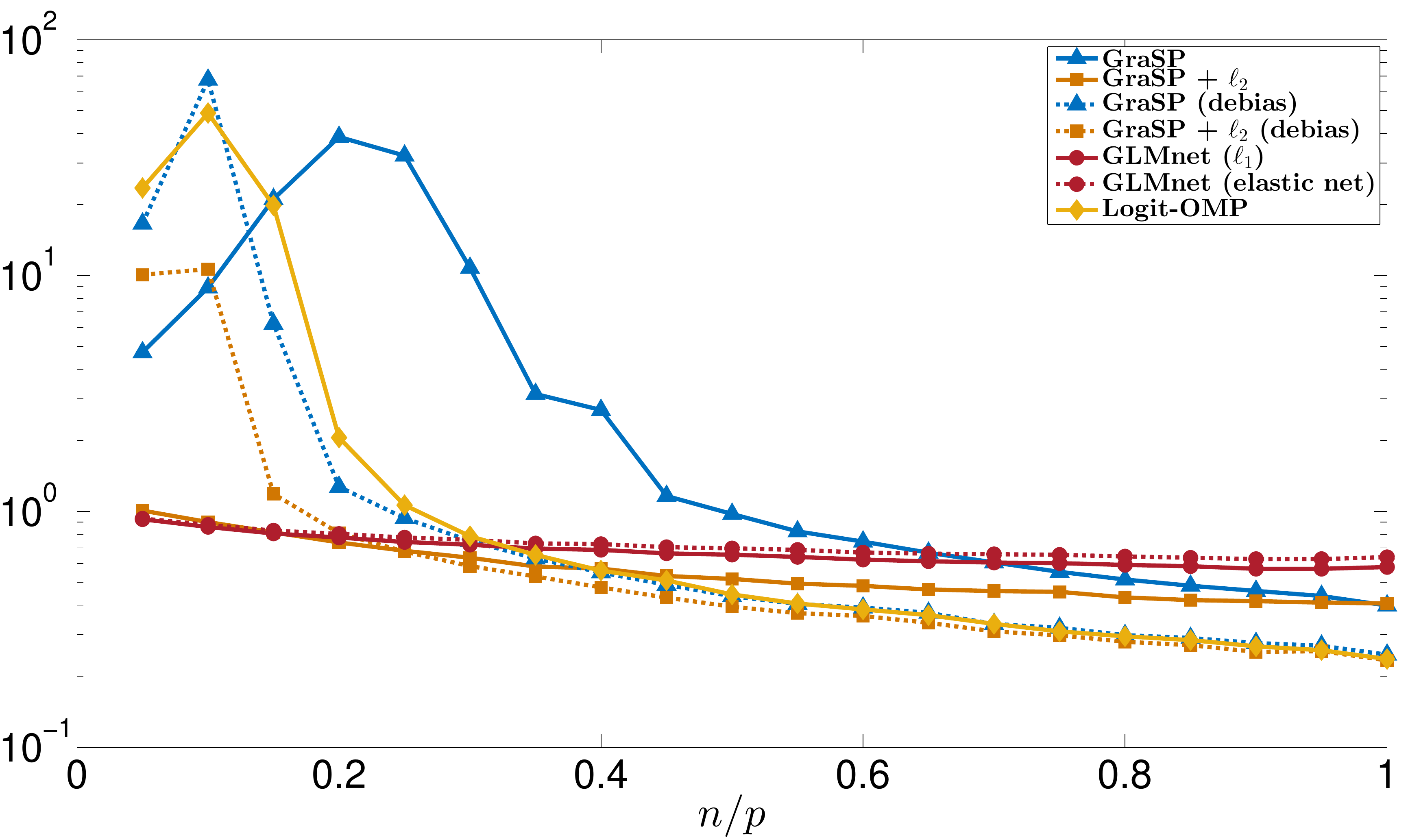}}
	\caption{Comparison of the average relative error (i.e., $\norm{\widehat{\vc{x}}-\vc{x}^\star}_2/\norm{\vc{x}^\star}_2$) in logarithmic scale at solutions obtained via GraSP, GraSP with $\ell_2$-penalty, LASSO, the elastic-net regularization, and Logit-OMP. The results of both GraSP methods with ``debiasing'' are also included.}\label{fig:experiments_RRE}
\end{figure}

Fig. \ref{fig:experiments_RRE} illustrates the performance of the same algorithms in terms of the relative error $\norm{\widehat{\vc{x}}-\vc{x^\star}}_2/\norm{\vc{x}^\star}_2$ where $\widehat{\vc{x}}$ denotes the estimate that the algorithms produce. Not surprisingly, none of the algorithms attain an arbitrarily small relative error. Furthermore, the parameter $\rho$ does not appear to affect the performance of the algorithms significantly. Without the $\ell_2$-regularization, at high sampling ratios GraSP provides an estimate that has a comparable error versus the $\ell_1$-regularization method. However, for mid to high sampling ratios both GraSP and GLMnet methods are outperformed by Logit-OMP. At low to mid sampling ratios, GraSP is unstable and does not converge to an estimate close to the true parameter. Logit-OMP shows similar behavior at lower sampling ratios. Performance of GraSP changes changes dramatically once we consider the $\ell_2$-regularization and/or the debiasing procedure. With $\ell_2$-regularization, GraSP achieves better relative error compared to GLMnet and ordinary GraSP for almost the entire range of tested sampling ratios. Applying the debiasing procedure has improved the performance of both GraSP methods except at very low sampling ratios. These variants of GraSP appear to perform better than Logit-OMP for almost the entire range of $n/p$.

\SBedit{
\subsection*{Real Data}

\begin{table}[t]
  \centering
  \caption{ARCENE}
    \begin{tabular}{llllllll}
    	\toprule
    	$s$ & G        & Gd       & G$\ell_2$ & G$\ell_2$d        & $\ell_1$ & E-net    & Logit-OMP          \\ \midrule
    	5   & 5.89E+01 & 5.75E-01 & 2.02E+01  & 5.24E-01          & 5.59E-01 & 6.43E-01 & \textbf{2.23E-01} \\
    	10  & 3.17E+02 & 5.43E-01 & 3.71E+01  & 4.53E-01          & 5.10E-01 & 5.98E-01 & \textbf{5.31E-07} \\
    	15  & 3.38E+02 & 6.40E-07 & 5.94E+00  & \textbf{1.42E-07} & 4.86E-01 & 5.29E-01 & 5.31E-07          \\
    	20  & 1.21E+02 & 3.44E-07 & 8.82E+00  & \textbf{3.08E-08} & 4.52E-01 & 5.19E-01 & 5.31E-07          \\
    	25  & 9.87E+02 & 1.13E-07 & 4.46E+01  & \textbf{1.35E-08} & 4.18E-01 & 4.96E-01 & 5.31E-07          \\ \bottomrule
    \end{tabular}%
  \label{tab:ARCENE}%
\end{table}%
\begin{table}
  \centering
  \caption{DEXTER}
    \begin{tabular}{llllllll}
    	\toprule
    	$s$ & G        & Gd                & G$\ell_2$ & G$\ell_2$d        & $\ell_1$ & E-net    & Logit-OMP          \\ \midrule
    	5   & 7.58E+00 & 3.28E-01          & 3.30E+00  & 2.80E-01          & 5.75E-01 & 6.08E-01 & \textbf{2.64E-01} \\
    	10  & 1.08E+00 & 1.79E-01          & 4.33E-01  & \textbf{1.28E-01} & 5.23E-01 & 5.33E-01 & 1.79E-01          \\
    	15  & 6.06E+00 & 1.71E-01          & 3.35E-01  & 1.17E-01          & 4.88E-01 & 4.98E-01 & \textbf{1.16E-01} \\
    	20  & 1.30E+00 & 8.84E-02          & 1.79E-01  & 8.19E-02          & 4.27E-01 & 4.36E-01 & \textbf{4.60E-02} \\
    	25  & 1.17E+00 & \textbf{2.51E-07} & 2.85E-01  & 1.17E-02          & 3.94E-01 & 4.12E-01 & 4.62E-03          \\
    	30  & 3.04E-01 & 5.83E-07          & 2.65E-01  & \textbf{1.77E-07} & 3.70E-01 & 3.88E-01 & 2.88E-07          \\
    	35  & 6.22E-01 & 2.08E-07          & 2.68E-01  & \textbf{1.19E-07} & 3.47E-01 & 3.72E-01 & 2.14E-07          \\
    	40  & 5.38E-01 & 2.01E-07          & 6.30E-02  & \textbf{1.27E-07} & 3.31E-01 & 3.56E-01 & 2.14E-07          \\
    	45  & 3.29E-01 & 2.11E-07          & 1.05E-01  & \textbf{1.47E-07} & 3.16E-01 & 3.41E-01 & 2.14E-07          \\
    	50  & 2.06E-01 & \textbf{1.31E-07} & 5.66E-02  & 1.46E-07          & 2.87E-01 & 3.11E-01 & 2.14E-07          \\
    	55  & 3.61E-02 & \textbf{1.20E-07} & 8.40E-02  & 1.31E-07          & 2.80E-01 & 2.89E-01 & 2.14E-07          \\
    	60  & 1.18E-01 & 2.46E-07          & 5.70E-02  & \textbf{1.09E-07} & 2.66E-01 & 2.82E-01 & 2.14E-07          \\
    	65  & 1.18E-01 & \textbf{7.86E-08} & 2.87E-02  & 9.47E-08          & 2.59E-01 & 2.75E-01 & 2.14E-07          \\
    	70  & 8.92E-02 & 1.17E-07          & 2.23E-02  & \textbf{8.15E-08} & 2.52E-01 & 2.69E-01 & 2.14E-07          \\
    	75  & 1.03E-01 & 8.54E-08          & 3.93E-02  & \textbf{7.94E-08} & 2.45E-01 & 2.69E-01 & 2.14E-07          \\ \bottomrule
    \end{tabular}%
  \label{tab:DEXTER}%
\end{table}%

We also conducted the same simulation on some of the data sets used in NIPS 2003 Workshop on feature extraction \citep{Guyon_Feature_2006}, namely the ARCENE and DEXTER data sets. The logistic loss values at obtained estimates are reported in Tables \ref{tab:ARCENE} and \ref{tab:DEXTER}. For each data set we applied the sparse logistic regression for a range of sparsity level $s$. The columns indicated by ``G'' correspond to different variants of GraSP. Suffixes $\ell_2$ and ``d'' indicate the $\ell_2$-regularization and the debiasing are applied, respectively. The columns indicated by $\ell_1$ and E-net correspond to the results of the $\ell_1$-regularization and the elastic-net regularization methods that are performed using the GLMnet package. The last column contains the result of the Logit-OMP algorithm.

The results for DEXTER data set show that GraSP variants without debiasing and the convex methods achieve comparable loss values in most cases, whereas the convex methods show significantly better performance on the ARCENE data set. Nevertheless, except for a few instances where Logit-OMP has the best performance, the smallest loss values in both data sets are attained by GraSP methods with debiasing step.}
\section{Discussion and Conclusion}\label{sec:conclusion}
In many applications understanding high dimensional data or systems
that involve these types of data can be reduced to identification of a
sparse parameter. For example, in gene selection problems researchers
are interested in locating a few genes among thousands of genes that
cause or contribute to a particular disease. These problems can
usually be cast as sparsity constrained optimizations. In this paper
we introduce a greedy algorithm called the Gradient Support
Pursuit(GraSP) as an approximate solver for a wide range of
sparsity-constrained optimization problems.

We provide theoretical convergence guarantees based on the notions of
a Stable Restricted Hessian (\SCond{}) for smooth cost functions and a
Stable Restricted Linearization (\NSCond{}) for non-smooth cost
functions, both of which are introduced in this paper. Our algorithm
generalizes the well-established sparse recovery algorithm CoSaMP that
merely applies in linear models with squared error loss. The \SCond{}
and \NSCond{} also generalize the well-known Restricted Isometry
Property for sparse recovery to the case of cost functions other than
the squared error. To provide a concrete example we studied the
requirements of GraSP for $\ell_2$-regularized logistic loss. Using a
similar approach one can verify \SCond{} condition for loss functions
that have Lipschitz-continuous gradient that incorporates a broad
family of loss functions.
 
 At medium- and large-scale problems computational cost of the GraSP
 algorithm is mostly affected by the inner convex optimization step
 whose complexity is polynomial in $s$. On the other hand, for very
 large-scale problems, especially with respect to the dimension of the
 input, $p$, the running time of the GraSP algorithm will be dominated
 by evaluation of the function and its gradient, whose computational
 cost grows with $p$. This problem is common in algorithms that only
 have deterministic steps; even ordinary coordinate-descent methods
 have this limitation \citep{nesterov_efficiency_2012}. Similar to
 improvements gained by using randomization in coordinate-descent
 methods \citep{nesterov_efficiency_2012}, introducing randomization
 in the GraSP algorithm could reduce its computational complexity at
 large-scale problems. This extension, however, is beyond the scope of
 this paper and we leave it for future work.

\bibliography{references}
\appendix
\section{Iteration Analysis For Smooth Cost Functions}\label{app:Smooth}
To analyze our algorithm we first establish a series of results on how
the algorithm operates on its current estimate, leading to an
iteration invariant property on the estimation error. Propositions \ref{pro:P1} and \ref{pro:P2} are used to prove Lemmas \ref{lem:lem1} and \ref{lem:lem2}. These Lemmas then are used to prove Lemma \ref{lem:lem3} that provides an iteration invariant which in turn yields the main result.

\begin{pro}
\label{pro:P1}Let $\mx{M}\left(t\right)$ be a matrix-valued function
such that for all $t\in\left[0,1\right]$, $\mx{M}\left(t\right)$ is symmetric
and its eigenvalues lie in interval $\left[B\left(t\right),A\left(t\right)\right]$
with $B\left(t\right)>0$. Then for any vector $\vc{v}$ we have \begin{align*}
\left(\intop_{0}^{1}B(t)\mathrm{d}t\right)\,\norm{\vc{v}}_{2}\leq\norm{\left(\intop_{0}^{1}\mx{M}(t)\mathrm{d}t\right)\,\vc{v}}_{2}\leq\left(\intop_{0}^{1}A(t)\mathrm{d}t\right)\,\norm{\vc{v}}_{2}.\end{align*}
\end{pro}
\begin{proof}
Let $\lambda_{\text{min}}\left(\cdot\right)$ and $\lambda_{\text{max}}\left(\cdot\right)$ denote the smallest and largest eigenvalue functions defined over the set of symmetric positive-definite
matrices, respectively. These functions are in order
concave and convex. Therefore, 
Jensen's inequality
yields
\begin{align*}
\lambda_{\text{min}}\left(\intop_{0}^{1}\mx{M}(t)\mathrm{d}t\right)\geq\intop_{0}^{1}\lambda_{\text{min}}\left(\mx{M}(t)\right)\mathrm{d}t\geq\intop_{0}^{1}B(t)\mathrm{d}t\end{align*}
 and \begin{align*}
\lambda_{\text{max}}\left(\intop_{0}^{1}\mx{M}(t)\mathrm{d}t\right) \leq\intop_{0}^{1}\lambda_{\text{max}}\left(\mx{M}(t)\right)\mathrm{d}t
\leq\intop_{0}^{1}A(t)\mathrm{d}t,\end{align*}
 which imply the desired result.\end{proof}
\begin{pro}
\label{pro:P2}Let $\mx{M}\left(t\right)$ be a matrix-valued function
such that for all $t\in\left[0,1\right]$ $\mx{M}\left(t\right)$ is symmetric
and its eigenvalues lie in interval $\left[B\left(t\right),A\left(t\right)\right]$
with $B\left(t\right)>0$. If $\Gamma$ is a subset of row/column indices of $\mx{M}\left(\cdot\right)$ then for any vector $\vc{v}$ we have \[
\norm{\left(\intop_{0}^{1}\mx{P}^{\mathrm{T}}_{\Gamma}\mx{M}(t)\mx{P}_{\Gamma^{c}}\mathrm{d}t\right)\,\vc{v}}_{2}\leq\intop_{0}^{1}\frac{A(t)-B\left(t\right)}{2}\mathrm{d}t\,\norm{\vc{v}}_{2}.\]
\end{pro}
\begin{proof}
Since $\mx{M}\left(t\right)$ is symmetric, it is also diagonalizable. Thus, for any vector
$\vc{v}$ we may write \[ B\left(t\right)\norm{\vc{v}}_{2}^{2}\leq
\vc{v}^{\mathrm{T}}\mx{M}\left(t\right)\vc{v}\leq
A\left(t\right)\norm{\vc{v}}_{2}^{2},\] and thereby \begin{align*}
 - \frac{A\left(t\right)\!-\!B\left(t\right)}{2}\leq\frac{\vc{v}^{\mathrm{T}}\left(\mx{M}\left(t\right)\!-\!\frac{A\left(t\right)\!+\!B\left(t\right)}{2}\mx{I}\right)\vc{v}}{\norm{\vc{v}}^{2}}  \leq\frac{A\left(t\right)\!-\!B\left(t\right)}{2}.\end{align*}
Since $\mx{M}\left(t\right)-\frac{A\left(t\right)+B\left(t\right)}{2}\mx{I}$ is also diagonalizable, it follows from the above inequality that $\norm{\mx{M}\left(t\right)-
\frac{A\left(t\right)+B\left(t\right)}{2}\mx{I}}
\leq\frac{A\left(t\right)-B\left(t\right)}{2}$.  Let
$\widetilde{\mx{M}}\left(t\right)=\mx{P}^{\mathrm{T}}_{\Gamma}\mx{M}\left(t\right)\mx{P}_{\Gamma^{c}}$.
Since $\widetilde{\mx{M}}\left(t\right)$ is a submatrix of
$\mx{M}\left(t\right)-\frac{A\left(t\right)+B\left(t\right)}{2}\mx{I}$ we
should have 
\begin{equation}
  \norm{\widetilde{\mx{M}}\left(t\right)}\leq\norm{\mx{M}\left(t\right)-\frac{A\left(t\right)+B\left(t\right)}{2}\mx{I}}\leq\frac{A\left(t\right)-B\left(t\right)}{2}.\label{eq:P2E1}\end{equation}
Finally, it follows from the convexity of the operator norm, Jensen's
inequality, and \eqref{eq:P2E1} that\begin{align*}
\norm{\intop_{0}^{1}\widetilde{\mx{M}}\left(t\right)\mathrm{d}t}\leq\intop_{0}^{1}\norm{\widetilde{\mx{M}}\left(t\right)}\mathrm{d}t
\leq\intop_{0}^{1}\frac{A(t)-B\left(t\right)}{2}\mathrm{d}t.\end{align*}
\end{proof}

To simplify notation we introduce functions\begin{align*}
\alpha_{k}\left(\vc{p},\vc{q}\right) & =\intop_{0}^{1}A_{k}\left(t\vc{q}+\left(1-t\right)\vc{p}\right)\mathrm{d}t\\
\beta_{k}\left(\vc{p},\vc{q}\right) & =\intop_{0}^{1}B_{k}\left(t\vc{q}+\left(1-t\right)\vc{p}\right)\mathrm{d}t\\
\gamma_{k}\left(\vc{p},\vc{q}\right) & =\alpha_{k}\left(\vc{p},\vc{q}\right)-\beta_{k}\left(\vc{p},\vc{q}\right),\end{align*}
 where $A_{k}\left(\cdot\right)$ and $B_{k}\left(\cdot\right)$ are
defined by \eqref{eq:E00} and \eqref{eq:E01}, respectively.
\begin{lem}
\label{lem:lem1} Let  $\st{R}$ denote the set $\mathrm{supp}\left(\widehat{\vc{x}}-\vc{x}^{\star}\right)$. The current estimate $\widehat{\vc{x}}$ then satisfies
\begin{align*}
\norm{\left(\widehat{\vc{x}}\!-\!\vc{x}^{\star}\!\right)|_{\st{Z}^{c}}}_{2}\leq & \frac{\gamma_{4s}\left(\widehat{\vc{x}},\!\vc{x}^{\star}\!\right)\!+\!\gamma_{2s}\left(\widehat{\vc{x}},\!\vc{x}^{\star}\!\right)}{2\beta_{2s}\left(\widehat{\vc{x}},\!\vc{x}^{\star}\!\right)}\norm{\widehat{\vc{x}}\!-\!\vc{x}^{\star}\!}_{2}\!+\!\frac{\norm{\nabla f\left(\vc{x}^{\star}\!\right)|_{\st{R}\!\backslash\st{Z}}}_{2}\!+\!\norm{\nabla f\left(\vc{x}^{\star}\!\right)|_{\st{Z}\!\backslash\! \st{R}}}_{2}}{\beta_{2s}\left(\widehat{\vc{x}},\!\vc{x}^{\star}\!\right)}.\end{align*}\end{lem}
\begin{proof} Since $\st{Z}=\text{supp}\left(\vc{z}_{2s}\right)$ and $\left|\st{R}\right|\leq 2s$
we have $\norm{\vc{z}|_{\st{R}}}_{2}\leq\norm{\vc{z}|_{\st{Z}}}_{2}$ and thereby 
\begin{align}
\norm{\vc{z}|_{\st{R}\backslash\st{Z}}}_{2}\leq\norm{\vc{z}|_{\st{Z}\backslash \st{R}}}_{2}.\label{eq:L1E0}
\end{align}
 Furthermore, because $\vc{z}=\nabla f\left(\widehat{\vc{x}}\right)$ we can write
\begin{align*}
\norm{\vc{z}|_{\st{R}\backslash\st{Z}}}_{2}\!\geq &
\norm{\nabla\!f\left(\widehat{\vc{x}}\right)\!|_{\st{R}\!\backslash\!\st{Z}}\!\!-\!\!\nabla\!
f\left(\vc{x}^{\star}\right)\!|_{\st{R}\!\backslash\!\st{Z}}}_{2}\!-\!\norm{\nabla\!f\left(\vc{x}^{\star}\right)\!|_{\st{R}\!\backslash\!\st{Z}}}_{2}\\ =
& \left\Vert\left(\intop_{0}^{1}\mx{P}^{\mathrm{T}}_{\st{R}\backslash\st{Z}}\mx{H}_{f}\left(t\widehat{\vc{x}}+\left(1-t\right)\vc{x}^{\star}\right)\mathrm{d}t\right) \left(\widehat{\vc{x}}-\vc{x}^{\star}\right)\right\Vert_{2}-\norm{\nabla
f\left(\vc{x}^{\star}\right)|_{\st{R}\backslash\st{Z}}}_{2}\\
\geq & \left\Vert
\left(\intop_{0}^{1}\mx{P}^{\mathrm{T}}_{\st{R}\backslash\st{Z}}\mx{H}_{f}\left(t\widehat{\vc{x}}+\left(1-t\right)\vc{x}^{\star}\right)\mx{P}_{\st{R}\backslash\st{Z}}\mathrm{d}t\right)\left(\widehat{\vc{x}}-\vc{x}^{\star}\right)|_{\st{R}\backslash\st{Z}}\right\Vert_{2}\!\!-\norm{\nabla f\left(\vc{x}^{\star}\right)|_{\st{R}\backslash\st{Z}}}_{2}\\ &
-\left\Vert
\left(\intop_{0}^{1}\mx{P}^{\mathrm{T}}_{\st{R}\backslash\st{Z}}\mx{H}_{f}\left(t\widehat{\vc{x}}+\left(1-t\right)\vc{x}^{\star}\right)\mx{P}_{\st{Z}\cap
  \st{R}}\mathrm{d}t\right)\left(\widehat{\vc{x}}-\vc{x}^{\star}\right)|_{\st{Z}\cap
  \st{R} }\right\Vert_{2}, \end{align*}
where we split the active coordinates (i.e., $\st{R}$) into the sets $\st{R}\backslash\st{Z}$ and $\st{Z}\cap\st{R}$ to apply the triangle inequality and obtain the last expression.   Applying Propositions \ref{pro:P1}
and \ref{pro:P2} yields
\begin{align}
  \norm{\vc{z}|_{\st{R}\!\backslash\st{Z}}}_{2}&\geq\!
  \beta_{2s}\left(\widehat{\vc{x}},\!\vc{x}^{\star}\!\right)\norm{\left(\widehat{\vc{x}}\!-\!\vc{x}^{\star}\!\right)|_{\st{R}\!\backslash\st{Z}}}_{2}\!-\!\frac{\gamma_{2s}\left(\widehat{\vc{x}},\!\vc{x}^{\star}\!\right)}{2}\norm{\left(\widehat{\vc{x}}\!-\!\vc{x}^{\star}\!\right)|_{\st{Z}\cap
    \st{R}}}_{2}\!-\!\norm{\nabla
  f\left(\vc{x}^{\star}\!\right)|_{\st{R}\!\backslash\st{Z}}}_{2}\nonumber\\
  &\geq\!
  \beta_{2s}\left(\widehat{\vc{x}},\!\vc{x}^{\star}\!\right)\norm{\left(\widehat{\vc{x}}\!-\!\vc{x}^{\star}\!\right)|_{\st{R}\backslash\st{Z}}}_{2}\!-\!  \frac{\gamma_{2s}\left(\widehat{\vc{x}},\!\vc{x}^{\star}\!\right)}{2}\norm{\widehat{\vc{x}}\!-\!\vc{x}^{\star}\!}_{2}
  \!-\!\norm{\nabla
  f\left(\vc{x}^{\star}\!\right)|_{\st{R}\backslash\st{Z}}}_{2}.\label{eq:L1E1}
  \end{align}
Similarly, we have \begin{align} \norm{\vc{z}|_{\st{Z}\backslash
    \st{R}}}_{2}\leq & \norm{\nabla
  f\left(\widehat{\vc{x}}\right)|_{\st{Z}\backslash \st{R}}-\nabla
  f\left(\vc{x}^{\star}\right)|_{\st{Z}\backslash \st{R}}}_{2}+\norm{\nabla f\left(\vc{x}^{\star}\right)|_{\st{Z}\backslash
    \st{R}}}_{2}\nonumber \\ = & \left\Vert
  \left(\intop_{0}^{1}\mx{P}^{\mathrm{T}}_{\st{Z}\backslash
    \st{R}}\mx{H}_{f}\left(t\widehat{\vc{x}}+\left(1-t\right)\vc{x}^{\star}\right)\mx{P}_{\st{R}}\mathrm{d}t\right)\left(\widehat{\vc{x}}-\vc{x}^{\star}\right)|_{\st{R}}\right\Vert_{2}+\norm{\nabla
  f\left(\vc{x}^{\star}\right)|_{\st{Z}\backslash \st{R}}}_{2}\nonumber
  \\ \leq &
  \frac{\gamma_{4s}\left(\widehat{\vc{x}},\vc{x}^{\star}\right)}{2}\norm{\left(\widehat{\vc{x}}-\vc{x}^{\star}\right)|_{\st{R}}}_{2}+\norm{\nabla
  f\left(\vc{x}^{\star}\right)|_{\st{Z}\backslash \st{R}}}_{2}\nonumber
  \\ =&
  \frac{\gamma_{4s}\left(\widehat{\vc{x}},\vc{x}^{\star}\right)}{2}\norm{\widehat{\vc{x}}-\vc{x}^{\star}}_{2}+\norm{\nabla
  f\left(\vc{x}^{\star}\right)|_{\st{Z}\backslash
    \st{R}}}_{2}.\label{eq:L1E2}\end{align} Combining \eqref{eq:L1E0},
\eqref{eq:L1E1}, and \eqref{eq:L1E2} we obtain
\begin{align*}
  \frac{\gamma_{4s}\!\left(\widehat{\vc{x}},\!\vc{x}^{\star}\!\right)}{2}\norm{\widehat{\vc{x}}\!-\!\vc{x}^{\star}}_{2}\!\!+\!\norm{\nabla f\!\left(\vc{x}^{\star}\right)|_{\st{Z}\!\backslash\!
    \st{R}}}_{2}&\geq\!\norm{\vc{z}|_{\st{Z}\backslash
    \st{R}}}_{2}\\
    & \geq\! \norm{\vc{z}|_{\st{R}\backslash\st{Z}}}_{2}\\
    & \geq\! \beta_{2s}\!\left(\widehat{\vc{x}},\!\vc{x}^{\star}\!\right)\norm{\left(\widehat{\vc{x}}\!-\!\vc{x}^{\star}\!\right)|_{\st{R}\!\backslash\!\st{Z}}}_{2}\!\!-\!\frac{\gamma_{2s}\!\left(\widehat{\vc{x}},\!\vc{x}^{\star}\!\right)}{2}\norm{\widehat{\vc{x}}\!-\!\vc{x}^{\star}}_{2} \\ & -\norm{\nabla
  f\!\left(\vc{x}^{\star}\right)|_{\st{R}\backslash\st{Z}}}_{2}.
  \end{align*} Since
$\st{R}=\text{supp}\left(\widehat{\vc{x}}\!-\!\vc{x}^{\star}\right)$, we have
$\norm{\left(\widehat{\vc{x}}\!-\!\vc{x}^{\star}\right)|_{\st{R}\backslash\st{Z}}}_{2}=\norm{\left(\widehat{\vc{x}}\!-\!\vc{x}^{\star}\right)|_{\st{Z}^{c}}}_{2}$.
Hence, \begin{align*}
  \norm{\left(\widehat{\vc{x}}\!-\!\vc{x}^{\star}\!\right)|_{\st{Z}^{c}}}_{2} &\leq\!
  \frac{\gamma_{4s}\left(\widehat{\vc{x}},\!\vc{x}^{\star}\!\right)\!+\!\gamma_{2s}\left(\widehat{\vc{x}},\!\vc{x}^{\star}\!\right)}{2\beta_{2s}\left(\widehat{\vc{x}},\!\vc{x}^{\star}\!\right)}\norm{\widehat{\vc{x}}\!-\!\vc{x}^{\star}\!}_{2}\!+\!\frac{\norm{\nabla
    f\left(\vc{x}^{\star}\!\right)|_{\st{R}\!\backslash\st{Z}}}_{2}\!+\!\norm{\nabla
    f\left(\vc{x}^{\star}\!\right)|_{\st{Z}\!\backslash\!
      \st{R}}}}{\beta_{2s}\left(\widehat{\vc{x}},\!\vc{x}^{\star}\!\right)}.\end{align*}
\end{proof}

\begin{lem}
\label{lem:lem2}The vector $\vc{b}$ given by \begin{align}
\vc{b}= & \arg\min f\left(\vc{x}\right)\ \mathrm{s.t.}\ \vc{x}|_{\st{T}^{c}}=0\label{eq:L2E0}\end{align}
satisfies \begin{align*}
\norm{\left.\vc{x}^{\star}\right\vert_\st{T}\!-\!\vc{b}}_{2}\le & \frac{\norm{\nabla f\left(\vc{x}^{\star}\right)|_{\st{T}}}_{2}}{\beta_{4s}\left(\vc{b},\vc{x}^{\star}\right)}\!
  +\!\frac{\gamma_{4s}\left(\vc{b},\vc{x}^{\star}\right)}{2\beta_{4s}\left(\vc{b},\vc{x}^{\star}\right)}\norm{\vc{x}^{\star}|_{\st{T}^{c}}}_{2}.\end{align*}
\end{lem}
\begin{proof}
We have \begin{align*}
\nabla f\left(\vc{x}^{\star}\right)\!-\!\nabla f\left(\vc{b}\right)=&\intop_{0}^{1}\mx{H}_{f}\left(t\vc{x}^{\star}\!+\!\left(1\!-\!t\right)\vc{b}\right)\mathrm{d}t\,\left(\vc{x}^{\star}\!-\!\vc{b}\right).\end{align*}
 Furthermore, since $\vc{b}$ is the solution to \eqref{eq:L2E0} we must
have $\nabla f\left(\vc{b}\right)|_{\st{T}}=0$. Therefore, \begin{align}
\nabla f\left(\vc{x}^{\star}\right)|_{\st{T}}= & \left(\intop_{0}^{1}\mx{P}^{\mathrm{T}}_{\st{T}}\mx{H}_{f}\left(t\vc{x}^{\star}+\left(1-t\right)\vc{b}\right)\mathrm{d}t\right)\,\left(\vc{x}^{\star}-\vc{b}\right)\nonumber \\
= & \left(\intop_{0}^{1}\mx{P}^{\mathrm{T}}_{\st{T}}\mx{H}_{f}\left(t\vc{x}^{\star}\!+\!\left(1\!-\!t\right)\vc{b}\right)\mx{P}_{\st{T}}\mathrm{d}t\right)\,\left(\vc{x}^{\star}-\vc{b}\right)|_{\st{T}}\nonumber \\
+ & \left(\intop_{0}^{1}\!\mx{P}^{\mathrm{T}}_{\st{T}}\mx{H}_{f}\!\left(t\vc{x}^{\star}\!+\!\left(1\!-\!t\right)\vc{b}\right)\mx{P}_{\st{T}^{c}}\mathrm{d}t\right)\,\left(\vc{x}^{\star}\!-\!\vc{b}\right)|_{\st{T}^{c}}.\label{eq:L2E1}\end{align}
 Since $f$ has $\mu_{4s}$-\SCond{} and $\left|\st{T}\cup\supp\left(t\vc{x}^\star+\left(1-t\right)\vc{b}\right)\right|\leq 4s$ for all $t\in\left[0,1\right]$, functions
$A_{4s}\left(\cdot\right)$ and $B_{4s}\left(\cdot\right)$, defined
using \eqref{eq:E2} and \eqref{eq:E3}, exist such that 
we have\begin{align*}
B_{4s}\left(t\vc{x}^{\star}+\left(1-t\right)\vc{b}\right) &\leq\lambda_{\text{min}}\!\left(\mx{P}^{\mathrm{T}}_{\st{T}}\mx{H}_{f}\left(t\vc{x}^{\star}\!+\!\left(1\!-\!t\right)\vc{b}\right)\mx{P}_{\st{T}}\right)\\
\intertext{and} A_{4s}\left(t\vc{x}^{\star}+\left(1-t\right)\vc{b}\right)&\geq \lambda_{\text{max}}\!\left(\mx{P}^{\mathrm{T}}_{\st{T}}\mx{H}_{f}\left(t\vc{x}^{\star}\!+\!\left(1\!-\!t\right)\vc{b}\right)\mx{P}_{\st{T}}\right).\end{align*}
 Thus, from Proposition \ref{pro:P1} we obtain \begin{align*}
\beta_{4s}\left(\vc{b},\vc{x}^{\star}\right) & \leq\lambda_{\text{min}}\left(\intop_{0}^{1}\mx{P}^{\mathrm{T}}_{\st{T}}\mx{H}_{f}\left(t\vc{x}^{\star}+\left(1-t\right)\vc{b}\right)\mx{P}_{\st{T}}\mathrm{d}t\right)\\
\intertext{and}
 \alpha_{4s}\left(\vc{b},\vc{x}^{\star}\right)&\geq \lambda_{\text{max}}\left(\intop_{0}^{1}\mx{P}^{\mathrm{T}}_{\st{T}}\mx{H}_{f}\left(t\vc{x}^{\star}+\left(1-t\right)\vc{b}\right)\mx{P}_{\st{T}}\mathrm{d}t\right).\end{align*}
 This result implies that the matrix $\intop_{0}^{1}\mx{P}^{\mathrm{T}}_{\st{T}}\mx{H}_{f}\left(t\vc{x}^{\star}+\left(1-t\right)\vc{b}\right)\mx{P}_{\st{T}}\mathrm{d}t$,
henceforth denoted by $\mx{W}$, is invertible and \begin{equation}
\frac{1}{\alpha_{4s}\left(\vc{b},\vc{x}^{\star}\right)}\leq\lambda_{\text{min}}\left(\mx{W}^{-1}\right)\leq\lambda_{\text{max}}\left(\mx{W}^{-1}\right)\leq\frac{1}{\beta_{4s}\left(\vc{b},\vc{x}^{\star}\right)},\label{eq:L2E2}\end{equation}
where we used the fact that $\lambda_{\max}\left(\mx{M}\right)\lambda_{\min}\left(\mx{M}^{-1}\right)=1$ for any positive-definite matrix $\mx{M}$, particularly for $\mx{W}$ and $\mx{W}^{-1}$. Therefore, by multiplying both sides of \eqref{eq:L2E1} by $\mx{W}^{-1}$ obtain \begin{align*}
\mx{W}^{-1}\nabla f\left(\vc{x}^{\star}\right)|_{\st{T}}= & \left(\vc{x}^{\star}-\vc{b}\right)|_{\st{T}}+ \mx{W}^{-1}\!\!\!\left(\intop_{0}^{1}\!\mx{P}^{\mathrm{T}}_{\st{T}}\mx{H}_{f}\!\left(t\vc{x}^{\star}\!+\!\left(1\!-\!t\right)\vc{b}\right) \mx{P}_{\st{T}^{c}}\mathrm{d}t\right)\,\vc{x}^{\star}|_{\st{T}^{c}},\end{align*}
where we also used the fact that $\left(\vc{x}^{\star}-\vc{b}\right)|_{\st{T}^{c}}=\vc{x}^{\star}|_{\st{T}^{c}}$. With $\st{S}^{\star}=\text{supp}\left(\vc{x}^{\star}\right)$, using triangle inequality, \eqref{eq:L2E2}, and Proposition \ref{pro:P2} then we obtain
\begin{align*}
\norm{\left.\vc{x}^{\star}\right\vert_\st{T}\!-\!\vc{b}}_{2}&=\norm{\left.\left(\vc{x}^{\star}\!\!-\!\vc{b}\right)\right\vert_{\st{T}}}_{2}\\ &\leq\!\norm{ \mx{W}^{-1}\!\!\!\left(\intop_{0}^{1}\!\!\mx{P}^{\mathrm{T}}_{\st{T}}\mx{H}_{f}\!\left(t\vc{x}^{\star}\!\!+\!\left(1\!-\!t\right)\vc{b}\right) \mx{P}_{\st{T}^{c}\cap \st{S}^{\star}}\mathrm{d}t\right)\,\vc{x}^{\star}|_{\st{T}^{c}\cap \st{S}^{\star}}\!}_{2}\!\!\!\!+\! \norm{\mx{W}^{-1}\nabla f\!\left(\vc{x}^{\star}\!\right)|_{\st{T}}\!}_{2}\\
 &\leq\! \frac{\norm{\nabla f\left(\vc{x}^{\star}\right)|_{\st{T}}}_{2}}{\beta_{4s}\left(\vc{b},\vc{x}^{\star}\right)}\!+\!\frac{\gamma_{4s}\left(\vc{b},\vc{x}^{\star}\right)}{2\beta_{4s}\left(\vc{b},\vc{x}^{\star}\right)}\norm{\vc{x}^{\star}|_{\st{T}^{c}}}_{2},\end{align*}
as desired.
\end{proof}
\begin{lem}[Iteration Invariant]
\label{lem:lem3} The estimation error in the current iteration,
$\norm{\widehat{\vc{x}}-\vc{x}^{\star}}_{2}$, and that in the next iteration,
$\norm{\vc{b}_{s}-\vc{x}^{\star}}_{2}$, are related by the
inequality:\textup{\begin{align*} \norm{\vc{b}_{s}-\vc{x}^{\star}}_{2}\leq &
    \frac{\gamma_{4s}\left(\widehat{\vc{x}},\vc{x}^{\star}\right)+\gamma_{2s}\left(\widehat{\vc{x}},\vc{x}^{\star}\right)}{2\beta_{2s}\left(\widehat{\vc{x}},\vc{x}^{\star}\right)}\left(1+\frac{\gamma_{4s}\left(\vc{b},\vc{x}^{\star}\right)}{\beta_{4s}\left(\vc{b},\vc{x}^{\star}\right)}\right)\norm{\widehat{\vc{x}}-\vc{x}^{\star}}_{2}\\ &
    +\left(1+\frac{\gamma_{4s}\left(\vc{b},\vc{x}^{\star}\right)}{\beta_{4s}\left(\vc{b},\vc{x}^{\star}\right)}\right)\frac{\norm{\nabla
      f\left(\vc{x}^{\star}\right)|_{\st{R}\backslash\st{Z}}}_{2}+\norm{\nabla
      f\left(\vc{x}^{\star}\right)|_{\st{Z}\backslash
        \st{R}}}_{2}}{\beta_{2s}\left(\widehat{\vc{x}},\vc{x}^{\star}\right)} +\frac{2\norm{\nabla
    f\left(\vc{x}^{\star}\right)|_{\st{T}}}_{2}}{\beta_{4s}\left(\vc{b},\vc{x}^{\star}\right)}.
    \end{align*}
}\end{lem}
\begin{proof}
Because $\st{Z}\subseteq\st{T}$ we must have $\st{T}^{c}\subseteq\st{Z}^{c}$. Therefore, we can write
$\norm{\vc{x}^{\star}|_{\st{T}^{c}}}_{2}=\norm{\left(\widehat{\vc{x}}-\vc{x}^{\star}\right)|_{\st{T}^{c}}}_{2}\leq\norm{\left(\widehat{\vc{x}}-\vc{x}^{\star}\right)|_{\st{Z}^{c}}}_{2}$.
Then using Lemma \ref{lem:lem1} we obtain \begin{align}
\norm{\vc{x}^{\star}|_{\st{T}^{c}}}_{2}\leq & \frac{\gamma_{4s}\!\left(\widehat{\vc{x}},\!\vc{x}^{\star}\right)\!+\!\gamma_{2s}\!\left(\widehat{\vc{x}},\!\vc{x}^{\star}\right)}{2\beta_{2s}\!\left(\widehat{\vc{x}},\!\vc{x}^{\star}\right)}\norm{\widehat{\vc{x}}\!-\!\vc{x}^{\star}}_{2} \!+\!\frac{\norm{\nabla f\!\left(\vc{x}^{\star}\right)|_{\st{R}\!\backslash\st{Z}}}_{2}\!+\!\norm{\nabla f\!\left(\vc{x}^{\star}\right)|_{\st{Z}\!\backslash\!\st{R}}}_{2}}{\beta_{2s}\!\left(\widehat{\vc{x}},\!\vc{x}^{\star}\right)}.\label{eq:L3E1}\end{align}
Furthermore, \begin{alignat}{2}
\norm{\vc{b}_{s}-\vc{x}^{\star}}_{2} & \leq \norm{\vc{b}_{s}-\left.\vc{x}^{\star}\right\vert_\st{T}}_{2} + \norm{\left.\vc{x}^\star\right\vert_{\st{T}^c}}_{2} & \nonumber \\  & \leq\norm{\left.\vc{x}^{\star}\right\vert_\st{T}-\vc{b}}_{2}+\norm{\vc{b}_{s}-\vc{b}}_{2}+\norm{\left.\vc{x}^\star\right\vert_{\st{T}^c}}_{2} & \leq 2\norm{\left.\vc{x}^{\star}\right\vert_\st{T}-\vc{b}}_{2} + \norm{\left.\vc{x}^\star\right\vert_{\st{T}^c}}_{2},\label{eq:hardThresh}
\end{alignat}
where the last inequality holds because $\norm{\left.\vc{x}^{\star}\right\vert_\st{T}}_{0}\leq s$ and $\vc{b}_{s}$ is the best $s$-term
approximation of $\vc{b}$. Therefore, using Lemma \ref{lem:lem2},
 \begin{align}
\norm{\vc{b}_{s}-\vc{x}^{\star}}_{2} & \leq\frac{2}{\beta_{4s}\left(\vc{b},\vc{x}^{\star}\right)}\norm{\nabla f\left(\vc{x}^{\star}\right)|_{\st{T}}}_{2} +\left(1+\frac{\gamma_{4s}\left(\vc{b},\vc{x}^{\star}\right)}{\beta_{4s}\left(\vc{b},\vc{x}^{\star}\right)}\right)\norm{\vc{x}^{\star}|_{\st{T}^{c}}}_{2}.\label{eq:L3E2}\end{align}
 Combining \eqref{eq:L3E1} and \eqref{eq:L3E2} we obtain \begin{align*}
\norm{\vc{b}_{s}\!-\!\vc{x}^{\star}}_{2}\leq & \frac{\gamma_{4s}\left(\widehat{\vc{x}},\vc{x}^{\star}\right)+\gamma_{2s}\left(\widehat{\vc{x}},\vc{x}^{\star}\right)}{2\beta_{2s}\left(\widehat{\vc{x}},\vc{x}^{\star}\right)}\left(1+\frac{\gamma_{4s}\left(\vc{b},\vc{x}^{\star}\right)}{\beta_{4s}\left(\vc{b},\vc{x}^{\star}\right)}\right)\norm{\widehat{\vc{x}}-\vc{x}^{\star}}_{2}\\
 & +\left(1+\frac{\gamma_{4s}\left(\vc{b},\vc{x}^{\star}\right)}{\beta_{4s}\left(\vc{b},\vc{x}^{\star}\right)}\right)\frac{\norm{\nabla f\left(\vc{x}^{\star}\right)|_{\st{R}\backslash\st{Z}}}_{2}+\norm{\nabla f\left(\vc{x}^{\star}\right)|_{\st{Z}\backslash \st{R}}}_{2}}{\beta_{2s}\left(\widehat{\vc{x}},\vc{x}^{\star}\right)}+\frac{2\norm{\nabla f\left(\vc{x}^{\star}\right)|_{\st{T}}}_{2}}{\beta_{4s}\left(\vc{b},\vc{x}^{\star}\right)}.\end{align*} \end{proof}
 
 Using the results above, we can now prove Theorem~\ref{thm:thm1}.

{\bf {\em Proof of Theorem} \ref{thm:thm1}.}
Using definition \ref{def:D1} it is easy to verify that for $k\leq k'$
and any vector $\vc{u}$ we have $A_{k}\left(\vc{u}\right)\leq A_{k'}\left(\vc{u}\right)$
and $B_{k}\left(\vc{u}\right)\geq B_{k'}\left(\vc{u}\right)$. Consequently,
for $k\leq k'$ and any pair of vectors $\vc{p}$ and $\vc{q}$ we have $\alpha_{k}\left(\vc{p},\vc{q}\right)\leq\alpha_{k'}\left(\vc{p},\vc{q}\right)$,
$\beta_{k}\left(\vc{p},\vc{q}\right)\geq\beta_{k'}\left(\vc{p},\vc{q}\right)$, and $\mu_{k}\leq\mu_{k'}$.
Furthermore, for any function that satisfies $\mu_{k}-$\SCond{} we can
write \begin{align*}
\frac{\alpha_{k}\left(\vc{p},\vc{q}\right)}{\beta_{k}\left(\vc{p},\vc{q}\right)} =\frac{\intop_{0}^{1}A_{k}\left(t\vc{q}+\left(1-t\right)\vc{p}\right)\mathrm{d}t}{\intop_{0}^{1}B_{k}\left(t\vc{q}+\left(1-t\right)\vc{p}\right)\mathrm{d}t}
 & \leq\frac{\intop_{0}^{1}\mu_{k}B_{k}\left(t\vc{q}+\left(1-t\right)\vc{p}\right)\mathrm{d}t}{\intop_{0}^{1}B_{k}\left(t\vc{q}+\left(1-t\right)\vc{p}\right)\mathrm{d}t}=\mu_{k},\end{align*}
 and thereby $\frac{\gamma_{k}\left(\vc{p},\vc{q}\right)}{\beta_{k}\left(\vc{p},\vc{q}\right)}\leq\mu_{k}-1$.
Therefore, applying Lemma \ref{lem:lem3} to the estimate in the $i$-th
iterate of the algorithm shows that \begin{align*}
\norm{\widehat{\vc{x}}^{\left(i\right)}\!-\!\vc{x}^{\star}}_{2}\leq & \left(\mu_{4s}-1\right)\mu_{4s}\norm{\widehat{\vc{x}}^{\left(i-1\right)}-\vc{x}^{\star}}_{2} +\frac{2\norm{\nabla f\left(\vc{x}^{\star}\right)|_{\st{T}}}_{2}}{\beta_{4s}\left(\vc{b},\vc{x}^{\star}\right)}\\ &+\!\mu_{4s} \frac{\norm{\nabla f\left(\vc{x}^{\star}\right)|_{\st{R}\backslash\st{Z}}}_{2}+\norm{\nabla f\left(\vc{x}^{\star}\right)|_{\st{Z}\backslash \st{R}}}_{2}}{\beta_{2s}\left(\widehat{\vc{x}}^{\left(i-1\right)},\vc{x}^{\star}\right)}\\
\leq & \left(\mu_{4s}^{2}-\mu_{4s}\right)\norm{\widehat{\vc{x}}^{\left(i-1\right)}-\vc{x}^{\star}}_{2}+ \frac{2}{\epsilon}\norm{\nabla f\left(\vc{x}^{\star}\right)|_{\mathcal{I}}}_{2}\!+\!\frac{2\mu_{4s}}{\epsilon}\norm{\nabla f\left(\vc{x}^{\star}\right)|_{\mathcal{I}}}_{2}\end{align*}
 Applying the assumption $\mu_{4s}\leq\frac{1+\sqrt{3}}{2}$ then yields \begin{align*}
\norm{\widehat{\vc{x}}^{\left(i\right)}\!-\!\vc{x}^{\star}}_{2} & \leq\frac{1}{2}\norm{\widehat{\vc{x}}^{\left(i-1\right)}\!-\!\vc{x}^{\star}}_{2}\!+\!\frac{3+\sqrt{3}}{\epsilon}\norm{\nabla f\left(\vc{x}^{\star}\right)|_{\mathcal{I}}}_{2}.\end{align*}
The theorem follows using this inequality recursively. \qquad \endproof

\section{Iteration Analysis For Non-Smooth Cost Functions}\label{app:Nonsmooth}
In this part we provide analysis of GraSP for non-smooth functions.
Definition \ref{def:D2} basically states that for any $k$-sparse vector $\vc{x}\in\mathbb{R}^{n}$,
$\alpha_{k}\left(\vc{x}\right)$ and $\beta_{k}\left(\vc{x}\right)$ are
in order the smallest and largest values for which 
\begin{align}
\beta_{k}\left(\vc{x}\right)\norm{\bsym{\Delta}}_{2}^{2}\leq\Breg f{\vc{x}+\bsym{\Delta}}{\vc{x}}\leq\alpha_{k}\left(\vc{x}\right)\norm{\bsym{\Delta}}_{2}^{2}\label{eq:LRSCP0}
\end{align}
 holds for all vectors $\bsym{\Delta}\in\mathbb{R}^{n}$ that satisfy $\left|\supp\left(\vc{x}\right)\cup\supp\left(\bsym{\Delta}\right)\right|\leq k$. By interchanging
$\vc{x}$ and $\vc{x}+\bsym{\Delta}$ in \eqref{eq:LRSCP0} and using the fact that
\begin{align*}
\mbox{\ensuremath{\Breg f{\vc{x}+\bsym{\Delta}}{\vc{x}}}+\ensuremath{\Breg f{\vc{x}}{\vc{x}+\bsym{\Delta}}}} & =\left\langle \nabla_{f}\left(\vc{x}+\bsym{\Delta}\right)-\nabla_{f}\left(\vc{x}\right),\bsym{\Delta}\right\rangle 
\end{align*}
 one can easily deduce 
\begin{align}
\left[\beta_{k}\!\left(\vc{x}\!+\!\bsym{\Delta}\right)\!+\!\beta_{k}\!\left(\vc{x}\right)\right]\!\norm{\bsym{\Delta}}_{2}^{2}&\!\leq\!\left\langle \nabla_{f}\!\left(\vc{x}\!+\!\bsym{\Delta}\right)\!-\!\nabla_{f}\!\left(\vc{x}\right),\bsym{\Delta}\right\rangle\!\leq\!\left[\alpha_{k}\!\left(\vc{x}\!+\!\bsym{\Delta}\right)\!+\!\alpha_{k}\!\left(\vc{x}\right)\right]\!\norm{\bsym{\Delta}}_{2}^{2}\!.\label{eq:LRSCP1}
\end{align}

Propositions \ref{pro:N1}, \ref{pro:N2}, and \ref{pro:N3} establish some basic inequalities regarding the restricted Bregman divergence under SRL assumption. Using these inequalities we prove Lemmas \ref{lem:N1} and \ref{lem:N2}. These two Lemmas are then used to prove an iteration invariant result in Lemma \ref{lem:N3} which in turn is used to prove Theorem \ref{thm:thm2}.

\begin{description}
\item [{Note}] In Propositions \ref{pro:N1}, \ref{pro:N2}, and \ref{pro:N3} we assume $\vc{x}_{1}$ and $\vc{x}_{2}$
are two vectors in $\mathbb{R}^{n}$ such that $\left\vert\supp\left(\vc{x}_1\right)\cup\supp\left(\vc{x}_2\right)\right\vert\leq r$.  Furthermore, we use the shorthand $\bsym{\Delta}=\vc{x}_{1}-\vc{x}_{2}$ and denote $\text{supp}\left(\bsym{\Delta}\right)$
by $\st{R}$. We also denote $\nabla_{f}\left(\vc{x}_{1}\right)-\nabla_{f}\left(\vc{x}_{2}\right)$
by $\bsym{\Delta}'$. To simplify the notation further the shorthands
$\bar{\alpha}_{l}$, $\bar{\beta}_{l}$, and $\bar{\gamma}_{l}$ are
used for $\bar{\alpha}_{l}\left(\vc{x}_{1},\vc{x}_{2}\right):=\alpha_{l}\left(\vc{x}_{1}\right)+\alpha_{l}\left(\vc{x}_{2}\right)$,
$\bar{\beta}_{l}\left(\vc{x}_{1},\vc{x}_{2}\right):=\beta_{l}\left(\vc{x}_{1}\right)+\beta_{l}\left(\vc{x}_{2}\right)$,
and $\bar{\gamma}_{l}\left(\vc{x}_{1},\vc{x}_{2}\right):=\bar{\alpha}_{l}\left(\vc{x}_{1},\vc{x}_{2}\right)-\bar{\beta}_{l}\left(\vc{x}_{1},\vc{x}_{2}\right)$,
respectively.
\end{description}
\begin{pro}\label{pro:N1}
Let $\st{R}'$ be a subset of $\st{R}$. Then the following inequalities
hold.
\begin{align}
\left|\bar{\alpha}_{r}\norm{\left.\bsym{\Delta}\right|_{\st{R}'}}_{2}^{2}-\left\langle \bsym{\Delta}',\left.\bsym{\Delta}\right|_{\st{R}'}\right\rangle \right| & \leq\bar{\gamma}_{r}\norm{\left.\bsym{\Delta}\right|_{\st{R}'}}_{2}\norm{\bsym{\Delta}}_{2}\label{eq:NP1ME1}\\
\left|\bar{\beta}_{r}\norm{\left.\bsym{\Delta}\right|_{\st{R}'}}_{2}^{2}-\left\langle \bsym{\Delta}',\left.\bsym{\Delta}\right|_{\st{R}'}\right\rangle \right| & \leq\bar{\gamma}_{r}\norm{\left.\bsym{\Delta}\right|_{\st{R}'}}_{2}\norm{\bsym{\Delta}}_{2}\label{eq:NP1ME2}
\end{align}
\end{pro}
\begin{proof}
Using \eqref{eq:LRSCP0} we can write 
\begin{align}
\beta_{r}\left(\vc{x}_{1}\right)\norm{\left.\bsym{\Delta}\right|_{\st{R}'}}_{2}^{2}t^{2}\leq\Breg f{\vc{x}_{1}-t\left.\bsym{\Delta}\right|_{\st{R}'}}{\vc{x}_{1}}\leq\alpha_{r}\left(\vc{x}_{1}\right)\norm{\left.\bsym{\Delta}\right|_{\st{R}'}}_{2}^{2}t^{2}\label{eq:NP1E1}\\
\beta_{r}\left(\vc{x}_{2}\right)\norm{\left.\bsym{\Delta}\right|_{\st{R}'}}_{2}^{2}t^{2}\leq\Breg f{\vc{x}_{2}-t\left.\bsym{\Delta}\right|_{\st{R}'}}{\vc{x}_{2}}\leq\alpha_{r}\left(\vc{x}_{2}\right)\norm{\left.\bsym{\Delta}\right|_{\st{R}'}}_{2}^{2}t^{2}\label{eq:NP1E2}
\end{align}
 and 
\begin{align}
\beta_{r}\left(\vc{x}_{1}\right)\norm{\bsym{\Delta}-t\left.\bsym{\Delta}\right|_{\st{R}'}}_{2}^{2}\leq\Breg f{\vc{x}_{2}+t\left.\bsym{\Delta}\right|_{\st{R}'}}{\vc{x}_{1}}\leq\alpha_{r}\left(\vc{x}_{1}\right)\norm{\bsym{\Delta}-t\left.\bsym{\Delta}\right|_{\st{R}'}}_{2}^{2}\label{eq:NP1E3}\\
\beta_{r}\left(\vc{x}_{2}\right)\norm{\bsym{\Delta}-t\left.\bsym{\Delta}\right|_{\st{R}'}}_{2}^{2}\leq\Breg f{\vc{x}_{1}-t\left.\bsym{\Delta}\right|_{\st{R}'}}{\vc{x}_{2}}\leq\alpha_{r}\left(\vc{x}_{2}\right)\norm{\bsym{\Delta}-t\left.\bsym{\Delta}\right|_{\st{R}'}}_{2}^{2},\label{eq:NP1E4}
\end{align}
 where $t$ is an arbitrary real number. Using the definition of the
Bregman divergence we can add \eqref{eq:NP1E1} and \eqref{eq:NP1E2}
to obtain 
\begin{align}
\bar{\beta}_{r}\norm{\left.\bsym{\Delta}\right|_{\st{R}'}}_{2}^{2}t^{2} & \leq f\left(\vc{x}_{1}-t\left.\bsym{\Delta}\right|_{\st{R}'}\right)-f\left(\vc{x}_{1}\right)+f\left(\vc{x}_{2}+t\left.\bsym{\Delta}\right|_{\st{R}'}\right)-f\left(\vc{x}_{2}\right)+\left\langle \bsym{\Delta}',\left.\bsym{\Delta}\right|_{\st{R}'}\right\rangle t\nonumber\\&\leq\bar{\alpha}_{r}\norm{\left.\bsym{\Delta}\right|_{\st{R}'}}_{2}^{2}t^{2}.\label{eq:NP1E5}
\end{align}
 Similarly, \eqref{eq:NP1E3} and \eqref{eq:NP1E4} yield 
\begin{align}
\bar{\beta}_{r}\norm{\bsym{\Delta}\!-\!t\!\left.\bsym{\Delta}\right|_{\st{R}'}}_{2}^{2} & \leq\! f\left(\vc{x}_{1}\!-\!t\!\left.\bsym{\Delta}\right|_{\st{R}'}\right)\!-\!f\left(\vc{x}_{1}\right)\!+\!f\left(\vc{x}_{2}+\!t\!\left.\bsym{\Delta}\right|_{\st{R}'}\right)\!-\!f\left(\vc{x}_{2}\right)\!+\!\left\langle \bsym{\Delta}',\bsym{\Delta}\!-\!t\!\left.\bsym{\Delta}\right|_{\st{R}'}\right\rangle \nonumber\\&\leq\bar{\alpha}_{r}\norm{\bsym{\Delta}-t\left.\bsym{\Delta}\right|_{\st{R}'}}_{2}^{2}.\label{eq:NP1E6}
\end{align}
 Expanding the quadratic bounds of \eqref{eq:NP1E6} and using \eqref{eq:NP1E5}
then we obtain 
\begin{align}
0 & \leq\bar{\gamma}_{r}\norm{\left.\bsym{\Delta}\right|_{\st{R}'}}_{2}^{2}t^{2}+2\left(\bar{\beta}_{r}\norm{\left.\bsym{\Delta}\right|_{\st{R}'}}_{2}^{2}-\left\langle \bsym{\Delta},\left.\bsym{\Delta}\right|_{\st{R}'}\right\rangle \right)t-\bar{\beta}_{r}\norm{\bsym{\Delta}}_{2}^{2}+\left\langle \bsym{\Delta}',\bsym{\Delta}\right\rangle \label{eq:NP1E7}\\
0 & \leq\bar{\gamma}_{r}\norm{\left.\bsym{\Delta}\right|_{\st{R}'}}_{2}^{2}t^{2}-2\left(\bar{\alpha}_{r}\norm{\left.\bsym{\Delta}\right|_{\st{R}'}}_{2}^{2}-\left\langle \bsym{\Delta},\left.\bsym{\Delta}\right|_{\st{R}'}\right\rangle \right)t+\bar{\alpha}_{r}\norm{\bsym{\Delta}}_{2}^{2}-\left\langle \bsym{\Delta}',\bsym{\Delta}\right\rangle .\label{eq:NP1E8}
\end{align}
 It follows from \eqref{eq:LRSCP1}, \eqref{eq:NP1E7}, and \eqref{eq:NP1E8}
that 
\begin{align}
0 & \leq\bar{\gamma}_{r}\norm{\left.\bsym{\Delta}\right|_{\st{R}'}}_{2}^{2}t^{2}+2\left(\bar{\beta}_{r}\norm{\left.\bsym{\Delta}\right|_{\st{R}'}}_{2}^{2}-\left\langle \bsym{\Delta},\left.\bsym{\Delta}\right|_{\st{R}'}\right\rangle \right)t+\bar{\gamma}_{r}\norm{\bsym{\Delta}}_{2}^{2}\label{eq:NP1E9}\\
0 & \leq\bar{\gamma}_{r}\norm{\left.\bsym{\Delta}\right|_{\st{R}'}}_{2}^{2}t^{2}-2\left(\bar{\alpha}_{r}\norm{\left.\bsym{\Delta}\right|_{\st{R}'}}_{2}^{2}-\left\langle \bsym{\Delta},\left.\bsym{\Delta}\right|_{\st{R}'}\right\rangle \right)t+\bar{\gamma}_{r}\norm{\bsym{\Delta}}_{2}^{2}.\label{eq:NP1E10}
\end{align}
 These two quadratic inequalities hold for any $t\in\mathbb{R}$ thus
their discriminants are not positive, i.e.,
\begin{align*}
\left(\bar{\beta}_{r}\norm{\left.\bsym{\Delta}\right|_{\st{R}'}}_{2}^{2}-\left\langle \bsym{\Delta}',\left.\bsym{\Delta}\right|_{\st{R}'}\right\rangle \right)^{2}-\bar{\gamma}_{r}^{2}\norm{\left.\bsym{\Delta}\right|_{\st{R}'}}_{2}^{2}\norm{\bsym{\Delta}}_{2}^{2} & \leq0\\
\left(\bar{\alpha}_{r}\norm{\left.\bsym{\Delta}\right|_{\st{R}'}}_{2}^{2}-\left\langle \bsym{\Delta}',\left.\bsym{\Delta}\right|_{\st{R}'}\right\rangle \right)^{2}-\bar{\gamma}_{r}^{2}\norm{\left.\bsym{\Delta}\right|_{\st{R}'}}_{2}^{2}\norm{\bsym{\Delta}}_{2}^{2} & \leq0
\end{align*}
 which yield the desired result.\end{proof}
\begin{pro}\label{pro:N2}
The following inequalities hold for $\st{R}'\subseteq\st{R}$.
\begin{align}
\left|\norm{\left.\bsym{\Delta}'\right|_{\st{R}'}}_{2}^{2}-\bar{\alpha}_{r}\left\langle \bsym{\Delta}',\left.\bsym{\Delta}\right|_{\st{R}'}\right\rangle \right| & \leq\bar{\gamma}_{r}\norm{\left.\bsym{\Delta}\right|_{\st{R}'}}_{2}\norm{\bsym{\Delta}}_{2}\label{eq:NP2ME1}\\
\left|\norm{\left.\bsym{\Delta}'\right|_{\st{R}'}}_{2}^{2}-\bar{\beta}_{r}\left\langle \bsym{\Delta}',\left.\bsym{\Delta}\right|_{\st{R}'}\right\rangle \right| & \leq\bar{\gamma}_{r}\norm{\left.\bsym{\Delta}\right|_{\st{R}'}}_{2}\norm{\bsym{\Delta}}_{2}\label{eq:NP2ME2}
\end{align}
\end{pro}
\begin{proof}
From \eqref{eq:LRSCP0} we have 
\begin{align}
\beta_{r}\left(\vc{x}_{1}\right)\norm{\left.\bsym{\Delta}'\right|_{\st{R}'}}_{2}^{2}t^{2}\leq\Breg f{\vc{x}_{1}-t\left.\bsym{\Delta}'\right|_{\st{R}'}}{\vc{x}_{1}}\leq\alpha_{r}\left(\vc{x}_{1}\right)\norm{\left.\bsym{\Delta}'\right|_{\st{R}'}}_{2}^{2}t^{2}\label{eq:NP2E1}\\
\beta_{r}\left(\vc{x}_{2}\right)\norm{\left.\bsym{\Delta}'\right|_{\st{R}'}}_{2}^{2}t^{2}\leq\Breg f{\vc{x}_{2}+t\left.\bsym{\Delta}'\right|_{\st{R}'}}{\vc{x}_{2}}\leq\alpha_{r}\left(\vc{x}_{2}\right)\norm{\left.\bsym{\Delta}'\right|_{\st{R}'}}_{2}^{2}t^{2}\label{eq:NP2E2}
\end{align}
 and 
\begin{align}
\beta_{r}\left(\vc{x}_{1}\right)\norm{\bsym{\Delta}-t\left.\bsym{\Delta}'\right|_{\st{R}'}}_{2}^{2}\leq\Breg f{\vc{x}_{2}+t\left.\bsym{\Delta}'\right|_{\st{R}'}}{\vc{x}_{1}}\leq\alpha_{r}\left(\vc{x}_{1}\right)\norm{\bsym{\Delta}-t\left.\bsym{\Delta}'\right|_{\st{R}'}}_{2}^{2}\label{eq:NP2E3}\\
\beta_{r}\left(\vc{x}_{2}\right)\norm{\bsym{\Delta}-t\left.\bsym{\Delta}'\right|_{\st{R}'}}_{2}^{2}\leq\Breg f{\vc{x}_{1}-t\left.\bsym{\Delta}'\right|_{\st{R}'}}{\vc{x}_{2}}\leq\alpha_{r}\left(\vc{x}_{2}\right)\norm{\bsym{\Delta}-t\left.\bsym{\Delta}'\right|_{\st{R}'}}_{2}^{2},\label{eq:NP2E4}
\end{align}
 for any $t\in\mathbb{R}$. By subtracting the sum of \eqref{eq:NP2E3}
and \eqref{eq:NP2E4} from that of \eqref{eq:NP2E1} and \eqref{eq:NP2E2}
we obtain 
\begin{align}
\bar{\beta}_{r}\norm{\left.\bsym{\Delta}'\right|_{\st{R}'}}_{2}^{2}t^{2}-\bar{\alpha}_{r}\norm{\bsym{\Delta}-t\left.\bsym{\Delta}'\right|_{\st{R}'}}_{2}^{2}&\leq2\left\langle \bsym{\Delta}',\left.\bsym{\Delta}'\right|_{\st{R}'}\right\rangle t-\left\langle \bsym{\Delta}',\bsym{\Delta}\right\rangle\nonumber\\& \leq\bar{\alpha}_{r}\norm{\left.\bsym{\Delta}'\right|_{\st{R}'}}_{2}^{2}t^{2}-\bar{\beta}_{r}\norm{\bsym{\Delta}-t\left.\bsym{\Delta}'\right|_{\st{R}'}}_{2}^{2}.\label{eq:NP2E5}
\end{align}
 Expanding the bounds of \eqref{eq:NP2E5} then yields 
\begin{align}
0\leq\bar{\gamma}_{r}\norm{\left.\bsym{\Delta}'\right|_{\st{R}'}}_{2}^{2}t^{2}+2\left(\left\langle \bsym{\Delta}',\left.\bsym{\Delta}'\right|_{\st{R}'}\right\rangle -\bar{\alpha}_{r}\left\langle \bsym{\Delta},\left.\bsym{\Delta}'\right|_{\st{R}'}\right\rangle \right)t+\bar{\alpha}_{r}\norm{\bsym{\Delta}}_{2}^{2}-\left\langle \bsym{\Delta'},\bsym{\Delta}\right\rangle \label{eq:NP2E6}\\
0\leq\bar{\gamma}_{r}\norm{\left.\bsym{\Delta}'\right|_{\st{R}'}}_{2}^{2}t^{2}-2\left(\left\langle \bsym{\Delta}',\left.\bsym{\Delta}'\right|_{\st{R}'}\right\rangle -\bar{\beta}_{r}\left\langle \bsym{\Delta},\left.\bsym{\Delta}'\right|_{\st{R}'}\right\rangle \right)t-\bar{\beta}_{r}\norm{\bsym{\Delta}}_{2}^{2}+\left\langle \bsym{\Delta'},\bsym{\Delta}\right\rangle .\label{eq:NP2E7}
\end{align}
 Note that $\left\langle \bsym{\Delta}',\left.\bsym{\Delta}'\right|_{\st{R}'}\right\rangle =\norm{\left.\bsym{\Delta}'\right|_{\st{R}'}}_{2}^{2}$
and $\left\langle \bsym{\Delta},\left.\bsym{\Delta}'\right|_{\st{R}'}\right\rangle =\left\langle \left.\bsym{\Delta}\right|_{\st{R}'},\bsym{\Delta}'\right\rangle $.
Therefore, using \eqref{eq:LRSCP1} we obtain 
\begin{align}
0\leq\bar{\gamma}_{r}\norm{\left.\bsym{\Delta}'\right|_{\st{R}'}}_{2}^{2}t^{2}+2\left(\norm{\left.\bsym{\Delta}'\right|_{\st{R}'}}_{2}^{2}-\bar{\alpha}_{r}\left\langle \bsym{\Delta}',\left.\bsym{\Delta}\right|_{\st{R}'}\right\rangle \right)t+\bar{\gamma}_{r}\norm{\bsym{\Delta}}_{2}^{2}\label{eq:NP2E8}\\
0\leq\bar{\gamma}_{r}\norm{\left.\bsym{\Delta}'\right|_{\st{R}'}}_{2}^{2}t^{2}-2\left(\norm{\left.\bsym{\Delta}'\right|_{\st{R}'}}_{2}^{2}-\bar{\beta}_{r}\left\langle \bsym{\Delta}',\left.\bsym{\Delta}\right|_{\st{R}'}\right\rangle \right)t+\bar{\gamma}_{r}\norm{\bsym{\Delta}}_{2}^{2}.\label{eq:NP2E9}
\end{align}
 Since the right-hand sides of \eqref{eq:NP2E8} and \eqref{eq:NP2E9} are
quadratics in $t$ and always non-negative for all values of $t\in\mathbb{R}$,
their discriminants cannot be positive. Thus we have 
\begin{align*}
\left(\norm{\left.\bsym{\Delta}'\right|_{\st{R}'}}_{2}^{2}-\bar{\alpha}_{r}\left\langle \bsym{\Delta}',\left.\bsym{\Delta}\right|_{\st{R}'}\right\rangle \right)^{2}-\bar{\gamma}_{r}^{2}\norm{\left.\bsym{\Delta}'\right|_{\st{R}'}}_{2}^{2}\norm{\bsym{\Delta}} ^{2} & \leq0\\
\left(\norm{\left.\bsym{\Delta}'\right|_{\st{R}'}}_{2}^{2}-\bar{\beta}_{r}\left\langle \bsym{\Delta}',\left.\bsym{\Delta}\right|_{\st{R}'}\right\rangle \right)^{2}-\bar{\gamma}_{r}^{2}\norm{\left.\bsym{\Delta}'\right|_{\st{R}'}}_{2}^{2}\norm{\bsym{\Delta}} ^{2} & \leq0,
\end{align*}
 which yield the desired result.\end{proof}
\begin{cor}\label{cor:N1}
The inequality 
\begin{align*}
\norm{\left.\bsym{\Delta}'\right|_{\st{R}'}}_{2} & \geq\bar{\beta}_{r}\norm{\left.\bsym{\Delta}\right|_{\st{R}'}}_{2}-\bar{\gamma}_{r}\norm{\left.\bsym{\Delta}\right|_{\st{R}\backslash\st{R}'}}_{2},
\end{align*}
 holds for $\st{R}'\subseteq\st{R}$.\end{cor}
\begin{proof}
It follows from \eqref{eq:NP2ME1} and \eqref{eq:NP1ME1} that 
\begin{align*}
-\!\norm{\!\left.\bsym{\Delta}'\right|_{\st{R}'}}_{2}^{2}\!+\!\bar{\alpha}_{r}^{2}\norm{\!\left.\bsym{\Delta}\right|_{\st{R}'}}_{2}^{2} & =\!-\!\norm{\!\left.\bsym{\Delta}'\right|_{\st{R}'}}_{2}^{2}\!+\!\bar{\alpha}_{r}\left\langle \bsym{\Delta}'\!,\!\left.\bsym{\Delta}\right|_{\st{R}'}\right\rangle\!+\!\bar{\alpha}_{r}\!\left[\bar{\alpha}_{r}\norm{\!\left.\bsym{\Delta}\right|_{\st{R}'}}_{2}^{2}\!-\!\left\langle \bsym{\Delta}'\!,\!\left.\bsym{\Delta}\right|_{\st{R}'}\right\rangle \right]\\
 & \leq\bar{\gamma}_{r}\norm{\left.\bsym{\Delta}'\right|_{\st{R}'}}_{2}\norm{\bsym{\Delta}}_{2}+\bar{\alpha}_{r}\bar{\gamma}_{r}\norm{\left.\bsym{\Delta}\right|_{\st{R}'}}_{2}\norm{\bsym{\Delta}}_{2}.
\end{align*}
 Therefore, after straightforward calculations we get 
\begin{align*}
\norm{\left.\bsym{\Delta}'\right|_{\st{R}'}}_{2} & \geq\frac{1}{2}\left(-\bar{\gamma}_{r}\norm{\bsym{\Delta}}_{2}+\left|2\bar{\alpha}_{r}\norm{\left.\bsym{\Delta}\right|_{\st{R}'}}_{2}-\bar{\gamma}_{r}\norm{\bsym{\Delta}}_{2}\right|\right)\\
 & \geq\bar{\alpha}_{r}\norm{\left.\bsym{\Delta}\right|_{\st{R}'}}_{2}-\bar{\gamma}_{r}\norm{\bsym{\Delta}}_{2}\\
 & \geq\bar{\beta}_{r}\norm{\left.\bsym{\Delta}\right|_{\st{R}'}}_{2}-\bar{\gamma}_{r}\norm{\left.\bsym{\Delta}\right|_{\st{R}\backslash\st{R}'}}_{2}.
\end{align*}
\end{proof}
\begin{pro}\label{pro:N3}
Suppose that $\st{K}$ is a subset of $\st{R}^{c}$ with at most $k$ elements.
Then we have 
\begin{align*}
\norm{\left.\bsym{\Delta}'\right|_{\st{K}}}_{2} & \leq\bar{\gamma}_{k+r}\norm{\bsym{\Delta}}_{2}.
\end{align*}
\end{pro}
\begin{proof}
Using \eqref{eq:LRSCP0} for any $t\in\mathbb{R}$ we can write 
\begin{align}
\beta_{k+r}\left(\vc{x}_{1}\right)\norm{\left.\bsym{\Delta}'\right|_{\st{K}}}_{2}^{2}t^{2}\leq\Breg f{\vc{x}_{1}+t\left.\bsym{\Delta}'\right|_{\st{K}}}{\vc{x}_{1}}\leq\alpha_{k+r}\left(\vc{x}_{1}\right)\norm{\left.\bsym{\Delta}'\right|_{\st{K}}}_{2}^{2}t^{2}\label{eq:NP3E1}\\
\beta_{k+r}\left(\vc{x}_{2}\right)\norm{\left.\bsym{\Delta}'\right|_{\st{K}}}_{2}^{2}t^{2}\leq\Breg f{\vc{x}_{2}-t\left.\bsym{\Delta}'\right|_{\st{K}}}{\vc{x}_{2}}\leq\alpha_{k+r}\left(\vc{x}_{2}\right)\norm{\left.\bsym{\Delta}'\right|_{\st{K}}}_{2}^{2}t^{2}\label{eq:NP3E2}
\end{align}
 and similarly 
\begin{align}
\beta_{k+r}\left(\vc{x}_{1}\right)\norm{\bsym{\Delta}\!+\!t\!\left.\bsym{\Delta}'\right|_{\st{K}}}_{2}^{2}\leq\Breg f{\vc{x}_{2}\!-\!t\left.\bsym{\Delta}'\right|_{\st{K}}}{\vc{x}_{1}}\leq\alpha_{k+r}\left(\vc{x}_{1}\right)\norm{\bsym{\Delta}\!+\!t\!\left.\bsym{\Delta}'\right|_{\st{K}}}_{2}^{2}\label{eq:NP3E3}\\
\beta_{k+r}\left(\vc{x}_{2}\right)\norm{\bsym{\Delta}\!+\!t\!\left.\bsym{\Delta}'\right|_{\st{K}}}_{2}^{2}\leq\Breg f{\vc{x}_{1}\!+\!t\!\left.\bsym{\Delta}'\right|_{\st{K}}}{\vc{x}_{2}}\leq\alpha_{k+r}\left(\vc{x}_{2}\right)\norm{\bsym{\Delta}\!+\!t\!\left.\bsym{\Delta}'\right|_{\st{K}}}_{2}^{2}.\label{eq:NP3E4}
\end{align}
 By subtracting the sum of \eqref{eq:NP3E3} and \eqref{eq:NP3E4} from
that of \eqref{eq:NP3E1} and \eqref{eq:NP3E2} we obtain 
\begin{align}
\bar{\beta}_{k+r}\norm{\left.\bsym{\Delta}'\right|_{\st{K}}}_{2}^{2}t^{2}-\bar{\alpha}_{k+r}\norm{\bsym{\Delta}\!+\! t\left.\bsym{\Delta}'\right|_{\st{K}}}_{2}^{2} & \leq-2t\left\langle \bsym{\Delta}',\left.\bsym{\Delta}'\right|_{\st{K}}\right\rangle -\left\langle \bsym{\Delta}',\bsym{\Delta}\right\rangle\nonumber\\& \leq\bar{\alpha}_{k+r}\norm{\left.\bsym{\Delta}'\right|_{\st{K}}}_{2}^{2}t^{2}-\bar{\beta}_{k+r}\norm{\bsym{\Delta}\!+\! t\left.\bsym{\Delta}'\right|_{\st{K}}}_{2}^{2}.\label{eq:NP3E5}
\end{align}
 Note that $\left\langle \bsym{\Delta}',\left.\bsym{\Delta}'\right|_{\st{K}}\right\rangle =\norm{\left.\bsym{\Delta}'\right|_{\st{K}}}_{2}^{2}$
and $\left\langle \bsym{\Delta},\left.\bsym{\Delta}'\right|_{\st{K}}\right\rangle =0$.
Therefore, \eqref{eq:LRSCP1} and \eqref{eq:NP3E5} imply 
\begin{align}
0 & \leq\bar{\gamma}_{k+r}\norm{\left.\bsym{\Delta}'\right|_{\st{K}}}_{2}^{2}t^{2}\pm2\norm{\left.\bsym{\Delta}'\right|_{\st{K}}}_{2}^{2}t+\bar{\gamma}_{k+r}\norm{\bsym{\Delta}}_{2}^{2}\label{eq:NP3E6}
\end{align}
hold for all $t\in\mathbb{R}.$ Hence, as quadratic functions of $t$,
the right-hand side of \eqref{eq:NP3E6} cannot have a positive discriminant.
Thus we must have 
\begin{align*}
\norm{\left.\bsym{\Delta}'\right|_{\st{K}}}_{2}^{4}-\bar{\gamma}_{k+r}^{2}\norm{\bsym{\Delta}}_{2}^{2}\norm{\left.\bsym{\Delta}'\right|_{\st{K}}}_{2}^{2} & \leq0
\end{align*}
 which yields the desired result.\end{proof}
\begin{lem}\label{lem:N1}
Let $\st{R}$ denote $\textnormal{supp}\left(\widehat{\vc{x}}-\vc{x}^{\star}\right)$.
Then we have \begin{align*}\norm{\left.\left(\widehat{\vc{x}}\!-\!\vc{x}^{\star}\right)\right|_{\st{Z}^{c}}}_{2}&\leq\!\frac{\bar{\gamma}_{2s}\left(\widehat{\vc{x}},\vc{x}^{\star}\right)\!+\!\bar{\gamma}_{4s}\left(\widehat{\vc{x}},\vc{x}^{\star}\!\right)}{\bar{\beta}_{2s}\left(\widehat{\vc{x}},\vc{x}^{\star}\!\right)}\norm{\widehat{\vc{x}}\!-\!\vc{x}^{\star}}_{2}\!+\!\frac{\norm{\left.\!\nabla_{f}\!\left(\vc{x}^{\star}\right)\right|_{\st{R}\!\backslash\!\st{Z}}}_{2}\!+\!\norm{\left.\!\nabla_{f}\!\left(\vc{x}^{\star}\right)\right|_{\st{Z}\!\backslash\!\st{R}}}_{2}}{\bar{\beta}_{2s}\left(\widehat{\vc{x}},\vc{x}^{\star}\right)}.
\end{align*}
\end{lem}
\begin{proof}
Given that $\st{Z}=\text{supp}\left(\vc{z}_{2s}\right)$ and $\left|\st{R}\right|\leq2s$
we have $\norm{\left.\vc{z}\right|_{\st{R}}}_{2}\leq\norm{\left.\vc{z}\right|_{\st{Z}}}_{2}$.
Hence 
\begin{align}
\norm{\left.\vc{z}\right|_{\st{R}\backslash\st{Z}}}_{2} & \leq\norm{\left.\vc{z}\right|_{\st{Z}\backslash\st{R}}}_{2}.\label{eq:NL1E1}
\end{align}
 Furthermore, using Corollary \ref{cor:N1} we can write 
\begin{align}
\norm{\left.\vc{z}\right|_{\st{R}\backslash\st{Z}}}_{2} & =\norm{\left.\nabla_{f}\left(\widehat{\vc{x}}\right)\right|_{\st{R}\backslash\st{Z}}}_{2}\nonumber \\
 & \geq\norm{\left.\left(\nabla_{f}\left(\widehat{\vc{x}}\right)-\nabla_{f}\left(\vc{x}^{\star}\right)\right)\right|_{\st{R}\backslash\st{Z}}}_{2}-\norm{\left.\nabla_{f}\left(\vc{x}^{\star}\right)\right|_{\st{R}\backslash\st{Z}}}_{2}\nonumber \\
 & \geq\!\bar{\beta}_{2s}\left(\widehat{\vc{x}},\!\vc{x}^{\star}\right)\norm{\left.\left(\widehat{\vc{x}}\!-\!\vc{x}^{\star}\right)\right|_{\st{R}\!\backslash\st{Z}}}_{2}\!\!-\bar{\gamma}_{2s}\left(\widehat{\vc{x}},\!\vc{x}^{\star}\right)\norm{\left.\left(\widehat{\vc{x}}\!-\!\vc{x}^{\star}\right)\right|_{\st{R}\cap\st{Z}}}_{2}\!-\!\norm{\left.\nabla_{f}\!\left(\vc{x}^{\star}\right)\right|_{\st{R}\!\backslash\st{Z}}}_{2}\nonumber \\
 & \geq\!\bar{\beta}_{2s}\left(\widehat{\vc{x}},\!\vc{x}^{\star}\right)\norm{\left.\left(\widehat{\vc{x}}\!-\!\vc{x}^{\star}\right)\right|_{\st{R}\backslash\st{Z}}}_{2}\!-\bar{\gamma}_{2s}\left(\widehat{\vc{x}},\!\vc{x}^{\star}\right)\norm{\widehat{\vc{x}}\!-\!\vc{x}^{\star}}_{2}\!-\!\norm{\left.\nabla_{f}\left(\vc{x}^{\star}\right)\right|_{\st{R}\!\backslash\st{Z}}}_{2}\!\!.\label{eq:NL1E2}
\end{align}
 Similarly, using Proposition \ref{pro:N3} we have 
\begin{align}
\norm{\left.\vc{z}\right|_{\st{Z}\backslash\st{R}}}_{2}=\norm{\left.\nabla_{f}\left(\widehat{\vc{x}}\right)\right|_{\st{Z}\backslash\st{R}}}_{2} & \leq\norm{\left.\left(\nabla_{f}\left(\widehat{\vc{x}}\right)-\nabla_{f}\left(\vc{x}^{\star}\right)\right)\right|_{\st{Z}\backslash\st{R}}}_{2}+\norm{\left.\nabla_{f}\left(\vc{x}^{\star}\right)\right|_{\st{Z}\backslash\st{R}}}_{2}\nonumber \\
 & \leq\bar{\gamma}_{4s}\left(\widehat{\vc{x}},\vc{x}^{\star}\right)\norm{\widehat{\vc{x}}-\vc{x}^{\star}}_{2}+\norm{\left.\nabla_{f}\left(\vc{x}^{\star}\right)\right|_{\st{Z}\backslash\st{R}}}_{2}.\label{eq:NL1E3}
\end{align}
 Combining \eqref{eq:NL1E1}, \eqref{eq:NL1E2}, and \eqref{eq:NL1E3}
then yields 
\begin{align*}
\bar{\gamma}_{4s}\left(\widehat{\vc{x}},\!\vc{x}^{\star}\right)\norm{\widehat{\vc{x}}\!-\!\vc{x}^{\star}}_{2}+\norm{\left.\nabla_{f}\left(\vc{x}^{\star}\right)\right|_{\st{Z}\!\backslash\st{R}}}_{2} & \geq -\bar{\gamma}_{2s}\left(\widehat{\vc{x}},\!\vc{x}^{\star}\right)\norm{\left.\left(\widehat{\vc{x}}\!-\!\vc{x}^{\star}\right)\right|_{\st{R}\cap\st{Z}}}_{2}\nonumber\\&+\!\bar{\beta}_{2s}\left(\widehat{\vc{x}},\!\vc{x}^{\star}\right)\norm{\left.\left(\widehat{\vc{x}}\!-\!\vc{x}^{\star}\right)\right|_{\st{R}\!\backslash\!\st{Z}}}_{2}\!-\!\norm{\left.\nabla_{f}\!\left(\vc{x}^{\star}\right)\right|_{\st{R}\!\backslash\!\st{Z}}}_{2}.
\end{align*}
 Note that $\left.\left(\widehat{\vc{x}}-\vc{x}^{\star}\right)\right|_{\st{R}\backslash\st{Z}}=\left.\left(\widehat{\vc{x}}-\vc{x}^{\star}\right)\right|_{\st{Z}^{c}}$.
Therefore, we have 
\begin{align*}
\norm{\left.\left(\widehat{\vc{x}}\!-\!\vc{x}^{\star}\right)\right|_{\st{Z}^{c}}}_{2}\leq\!\frac{\bar{\gamma}_{2s}\left(\widehat{\vc{x}},\!\vc{x}^{\star}\right)\!+\!\bar{\gamma}_{4s}\left(\widehat{\vc{x}},\!\vc{x}^{\star}\right)}{\bar{\beta}_{2s}\left(\widehat{\vc{x}},\!\vc{x}^{\star}\right)}\norm{\widehat{\vc{x}}\!-\!\vc{x}^{\star}}_{2}\!+\!\frac{\norm{\left.\nabla_{f}\!\left(\vc{x}^{\star}\right)\right|_{\st{R}\!\backslash\st{Z}}}_{2}\!+\!\norm{\left.\nabla_{f}\!\left(\vc{x}^{\star}\right)\right|_{\st{Z}\!\backslash\!\st{R}}}_{2}}{\bar{\beta}_{2s}\left(\widehat{\vc{x}},\!\vc{x}^{\star}\right)}.
\end{align*}
\end{proof}
\begin{lem}\label{lem:N2}
The vector $\vc{b}$ given by 
\begin{align}
\vc{b}=\arg\min_\vc{x}\ f\left(\vc{x}\right)\quad\mathrm{s.t.}\ \vc{x}|_{\st{T}^{c}}=\mathbf{0}\label{eq:NL2ME1}
\end{align}
 satisfies $\norm{\left.\vc{x}^{\star}\right\vert_\st{T}-\vc{b}}_{2}\leq\frac{\norm{\left.\nabla_{f}\left(\vc{x}^{\star}\right)\right|_{\st{T}}}_{2}}{\bar{\beta}_{4s}\left(\vc{x}^{\star},\vc{b}\right)}+\left(1+\frac{\bar{\gamma}_{4s}\left(\vc{x}^{\star},\vc{b}\right)}{\bar{\beta}_{4s}\left(\vc{x}^{\star},\vc{b}\right)}\right)\norm{\left.\vc{x}^{\star}\right|_{\st{T}^{c}}}_{2}$.\end{lem}
\begin{proof}
Since $\vc{b}$ satisfies \eqref{eq:NL2ME1} we must have $\left.\nabla_{f}\left(\vc{b}\right)\right|_{\st{T}}=\mathbf{0}$.
Then it follows from Corollary \ref{cor:N1} that
\begin{align*}
\norm{\left.\vc{x}^{\star}\right\vert_\st{T}-\vc{b}}_{2} &= \norm{\left.\left(\vc{x}^{\star}-\vc{b}\right)\right\vert_\st{T}}_{2}\\
& \leq\frac{\norm{\left.\nabla_{f}\left(\vc{x}^{\star}\right)\right|_{\st{T}}}_{2}}{\bar{\beta}_{4s}\left(\vc{x}^{\star},\vc{b}\right)}+\frac{\bar{\gamma}_{4s}\left(\vc{x}^{\star},\vc{b}\right)}{\bar{\beta}_{4s}\left(\vc{x}^{\star},\vc{b}\right)}\norm{\left.\vc{x}^{\star}\right|_{\st{T}^{c}}}_{2}.
\end{align*}
\end{proof}
\begin{lem}\label{lem:N3}
The estimation error of the current iterate \textnormal{(}i.e., $\norm{\widehat{\vc{x}}-\vc{x}^\star}_{2}$\textnormal{)} and that of the next iterate \textnormal{(}i.e., $\norm{\vc{b}_{s}-\vc{x}^\star}_{2}$\textnormal{)} are related by the inequality: 
\begin{align*}
\norm{\vc{b}_{s}\!-\!\vc{x}^{\star}}_{2} & \leq\left(1\!+\!\frac{2\bar{\gamma}_{4s}\left(\vc{x}^{\star},\vc{b}\right)}{\bar{\beta}_{4s}\left(\vc{x}^{\star},\vc{b}\right)}\right)\frac{\bar{\gamma}_{2s}\left(\widehat{\vc{x}},\vc{x}^{\star}\right)\!+\!\bar{\gamma}_{4s}\left(\widehat{\vc{x}},\vc{x}^{\star}\right)}{\bar{\beta}_{2s}\left(\widehat{\vc{x}}^{i},\vc{x}^{\star}\right)}\norm{\widehat{\vc{x}}\!-\!\vc{x}^{\star}}_{2}\!+\!\frac{2\norm{\left.\nabla_{f}\left(\vc{x}^{\star}\right)\right|_{\st{T}}}_{2}}{\bar{\beta}_{4s}\left(\vc{x}^{\star},\vc{b}\right)}\\
 & \!+\!\left(1\!+\!\frac{\bar{2\gamma}_{4s}\left(\vc{x}^{\star},\vc{b}\right)}{\bar{\beta}_{4s}\left(\vc{x}^{\star},\vc{b}\right)}\right)\frac{\norm{\left.\nabla_{f}\left(\vc{x}^{\star}\right)\right|_{\st{R}\backslash\st{Z}}}_{2}\!+\!\norm{\left.\nabla_{f}\left(\vc{x}^{\star}\right)\right|_{\st{Z}\backslash\st{R}}}_{2}}{\bar{\beta}_{2s}\left(\widehat{\vc{x}},\vc{x}^{\star}\right)}.
\end{align*}
\end{lem}

\begin{proof}
Since $\st{T}^{c}\subseteq\st{Z}^{c}$ we have $\norm{\left.\vc{x}^{\star}\right|_{\st{T}^{c}}}_{2}=\norm{\left.\left(\widehat{\vc{x}}-\vc{x}^{\star}\right)\right|_{\st{T}^{c}}}_{2}\leq\norm{\left.\left(\widehat{\vc{x}}-\vc{x}^{\star}\right)\right|_{\st{Z}^{c}}}_{2}$.
Therefore, applying Lemma \ref{lem:N1} yields 
\begin{align}
\norm{\left.\vc{x}^{\star}\right|_{\st{T}^{c}}}_{2} & \leq\!\frac{\bar{\gamma}_{2s}\!\left(\widehat{\vc{x}},\!\vc{x}^{\star}\right)\!+\!\bar{\gamma}_{4s}\!\left(\widehat{\vc{x}},\!\vc{x}^{\star}\right)}{\bar{\beta}_{2s}\!\left(\widehat{\vc{x}},\!\vc{x}^{\star}\right)}\norm{\widehat{\vc{x}}\!-\!\vc{x}^{\star}}_{2}\!+\!\frac{\norm{\left.\nabla_{f}\!\left(\vc{x}^{\star}\right)\right|_{\st{R}\!\backslash\!\st{Z}}}_{2}\!+\!\norm{\left.\nabla_{f}\!\left(\vc{x}^{\star}\right)\right|_{\st{Z}\!\backslash\!\st{R}}}_{2}}{\bar{\beta}_{2s}\!\left(\widehat{\vc{x}},\!\vc{x}^{\star}\right)}.\label{eq:NL3E1}
\end{align}
 Furthermore, as showed by \eqref{eq:hardThresh} during the proof of Lemma \ref{lem:lem3}, we again have
 \begin{alignat*}{1}
\norm{\vc{b}_{s}-\vc{x}^{\star}}_{2} & \leq 2\norm{\left.\vc{x}^{\star}\right\vert_\st{T}-\vc{b}}_{2} + \norm{\left.\vc{x}^\star\right\vert_{\st{T}^c}}_{2}.
\end{alignat*}
 Hence, it follows from
Lemma \ref{lem:N2} that 
\begin{align}
\norm{\vc{b}_{s}-\vc{x}^{\star}}_{2}\leq\frac{2\norm{\left.\nabla_{f}\left(\vc{x}^{\star}\right)\right|_{\st{T}}}_{2}}{\bar{\beta}_{4s}\left(\vc{x}^{\star},\vc{b}\right)}+\left(1+\frac{2\bar{\gamma}_{4s}\left(\vc{x}^{\star},\vc{b}\right)}{\bar{\beta}_{4s}\left(\vc{x}^{\star},\vc{b}\right)}\right)\norm{\left.\vc{x}^{\star}\right|_{\st{T}^{c}}}_{2}.\label{eq:NL3E2}
\end{align}
 Combining \eqref{eq:NL3E1} and \eqref{eq:NL3E2} yields 
\begin{align*}
\norm{\vc{b}_{s}-\vc{x}^{\star}}_{2} & \leq \left(1\!+\!\frac{2\bar{\gamma}_{4s}\left(\vc{x}^{\star},\vc{b}\right)}{\bar{\beta}_{4s}\left(\vc{x}^{\star},\vc{b}\right)}\right)\frac{\bar{\gamma}_{2s}\left(\widehat{\vc{x}},\vc{x}^{\star}\right)\!+\!\bar{\gamma}_{4s}\left(\widehat{\vc{x}},\vc{x}^{\star}\right)}{\bar{\beta}_{2s}\left(\widehat{\vc{x}},\vc{x}^{\star}\right)}\norm{\widehat{\vc{x}}-\vc{x}^{\star}}_{2}\!+\!\frac{2\norm{\left.\nabla_{f}\left(\vc{x}^{\star}\right)\right|_{\st{T}}}_{2}}{\bar{\beta}_{4s}\left(\vc{x}^{\star},\vc{b}\right)}\\
 & \!+\!\left(1\!+\!\frac{2\bar{\gamma}_{4s}\left(\vc{x}^{\star},\vc{b}\right)}{\bar{\beta}_{4s}\left(\vc{x}^{\star},\vc{b}\right)}\right)\frac{\norm{\left.\nabla_{f}\left(\vc{x}^{\star}\right)\right|_{\st{R}\backslash\st{Z}}}_{2}\!+\!\norm{\left.\nabla_{f}\left(\vc{x}^{\star}\right)\right|_{\st{Z}\backslash\st{R}}}_{2}}{\bar{\beta}_{2s}\left(\widehat{\vc{x}},\vc{x}^{\star}\right)}.
\end{align*}
\end{proof}

{\bf {\em Proof of Theorem} \ref{thm:thm2}.}
Let the vectors involved in the $j$-th iteration of the algorithm be denoted by superscript $\left(j\right)$. Given that $\mu_{4s}\leq\frac{3+\sqrt{3}}{4}$ we have 
\begin{alignat*}{3}
\frac{\bar{\gamma}_{4s}\left(\widehat{\vc{x}}^{\left(j\right)},\vc{x}^{\star}\right)}{\bar{\beta}_{4s}\left(\widehat{\vc{x}}^{\left(j\right)},\vc{x}^{\star}\right)}&\leq \frac{\sqrt{3}-1}{4}&\quad\text{and}\quad& 1+\frac{\bar{2\gamma}_{4s}\left(\vc{x}^{\star},\vc{b}^{\left(j\right)}\right)}{\bar{\beta}_{4s}\left(\vc{x}^{\star},\vc{b}^{\left(j\right)}\right)}&\leq\frac{1+\sqrt{3}}{2}, 
\end{alignat*}
that yield, 
\begin{align*}
\left(1+\frac{2\bar{\gamma}_{4s}\left(\vc{x}^{\star},\vc{b}\right)}{\bar{\beta}_{4s}\left(\vc{x}^{\star},\vc{b}\right)}\right)\frac{\bar{\gamma}_{2s}\left(\widehat{\vc{x}}^{\left(j\right)},\vc{x}^{\star}\right)+\bar{\gamma}_{4s}\left(\widehat{\vc{x}}^{\left(j\right)},\vc{x}^{\star}\right)}{\bar{\beta}_{2s}\left(\widehat{\vc{x}}^{\left(j\right)},\vc{x}^{\star}\right)} & \leq\frac{1+\sqrt{3}}{2}\times\frac{2\bar{\gamma}_{4s}\left(\widehat{\vc{x}}^{\left(j\right)},\vc{x}^{\star}\right)}{\bar{\beta}_{4s}\left(\widehat{\vc{x}}^{\left(j\right)},\vc{x}^{\star}\right)}\\
 & \leq\frac{1+\sqrt{3}}{2}\times\frac{\sqrt{3}-1}{2}\\
 & =\frac{1}{2}.
\end{align*}
 Therefore, it follows from Lemma \ref{lem:N3} that 
\begin{align*}
\norm{\widehat{\vc{x}}^{^{\left(j+1\right)}}-\vc{x}^{\star}}_{2} & \leq\frac{1}{2}\norm{\widehat{\vc{x}}^{\left(j\right)}-\vc{x}^{\star}}_{2}+\frac{3+\sqrt{3}}{\epsilon}\norm{\left.\nabla_{f}\left(\vc{x}^{\star}\right)\right|_{\st{I}}}_{2}.
\end{align*}
 Applying this inequality recursively for $j=0,1,\cdots,i-1$ then yields 
\begin{align*}
\norm{\widehat{\vc{x}}-\vc{x}^{\star}}_{2}\leq2^{-i}\norm{\vc{x}^{\star}}_{2}+\frac{6+2\sqrt{3}}{\epsilon}\norm{\left.\nabla_{f}\left(\vc{x}^{\star}\right)\right|_{\st{I}}}_{2}
\end{align*}
 which is the the desired result.\qquad\endproof

\end{document}